\documentclass{article}

\PassOptionsToPackage{numbers, compress}{natbib}
\usepackage[dandb, final]{neurips_2025}

\usepackage{arydshln}
\usepackage{amsmath}
\usepackage[utf8]{inputenc} % allow utf-8 input
\usepackage[T1]{fontenc}    % use 8-bit T1 fonts
\usepackage{hyperref}       % hyperlinks
\usepackage{url}            % simple URL typesetting
\usepackage{booktabs}       % professional-quality tables
\usepackage{amsfonts}       % blackboard math symbols
\usepackage{nicefrac}       % compact symbols for 1/2, etc.
\usepackage{microtype}      % microtypography
\usepackage[usenames,dvipsnames]{xcolor}         % colors
\usepackage{appendix}
\usepackage{graphicx}
\usepackage{booktabs}   % For better horizontal rules
\usepackage{multirow}   % For cells spanning multiple rows
\usepackage{array}      % Often useful with multirow, though maybe not strictly needed here
\usepackage{geometry}   % Optional: Adjust margins if the table 
\usepackage{multirow}
\usepackage{colortbl}
\usepackage{wrapfig}
\usepackage{siunitx} 
\usepackage[T1]{fontenc}
\usepackage{enumitem}
\usepackage{comicneue}

\definecolor{lightblue}{rgb}{0.7, 0.85, 0.9}
\definecolor{lightyellow}{rgb}{0.98, 0.98, 0.82} 
\definecolor{lightred}{rgb}{1.0, 0.8, 0.8}
\definecolor{lightgreen}{rgb}{0.8, 1.0, 0.8}
\definecolor{Gray}{gray}{0.94}
\definecolor{lightpurple}{rgb}{0.9, 0.8, 0.9}
\definecolor{MyDarkOlive}{HTML}{747106}

\usepackage{amsmath} % For math environments like align* and formulas
\usepackage{amssymb} % For symbols like \mathbb
\usepackage{algorithm} % For the algorithm environment
\usepackage{algpseudocode} % For pseudocode commands like \State, \For, \EndFor

\usepackage{caption} % For table captions

\definecolor{red}{rgb}{1,0,0} 

\algrenewcommand\algorithmicrequire{\textbf{Inputs:}}
\algrenewcommand\algorithmicensure{\textbf{Outputs:}}
\title{MultiHuman-Testbench: Benchmarking Image Generation for Multiple Humans}

\author{%
  Shubhankar Borse$^\S$ \quad Seokeon Choi \quad Sunghyun Park \quad Jeongho Kim \quad Shreya Kadambi \\
  \textbf{Risheek Garrepalli \quad Sungrack Yun \quad Munawar Hayat$^\S$ \quad Fatih Porikli} \\
  Qualcomm AI Research\thanks{Qualcomm AI Research is an initiative of Qualcomm Technologies, Inc.}\\
  $^\S$\texttt{\{sborse, mhayat\}@qti.qualcomm.com}\\
  }
\begin{document}

\maketitle

\begin{abstract}
  Generation of images containing multiple humans, performing complex actions, while preserving their facial identities, is a significant challenge. A major factor contributing to this is the lack of a dedicated benchmark. To address this, we introduce MultiHuman-Testbench, a novel benchmark for rigorously evaluating generative models for multi-human generation. The benchmark comprises 1,800 samples, including carefully curated text prompts, describing a range of simple to complex human actions. These prompts are matched with a total of 5,550 unique human face images, sampled uniformly to ensure diversity across  age, ethnic background, and gender. Alongside captions, we provide human-selected pose conditioning images which accurately match the prompt. We propose a multi-faceted evaluation suite employing four key metrics to quantify face count, ID similarity, prompt alignment, and action detection. We conduct a thorough evaluation of a diverse set of models, including zero-shot approaches and training-based methods, with and without regional priors. We also propose novel techniques to incorporate image and region isolation using human segmentation and Hungarian matching, significantly improving ID similarity. Our proposed benchmark and key findings provide valuable insights and a standardized tool for advancing research in multi-human image generation. The dataset and evaluation codes will be available at \texttt{https://github.com/Qualcomm-AI-research/MultiHuman-Testbench}.
  
\end{abstract}

\vspace{-15pt}
\section{Introduction}
\vspace{-10pt}
\begin{figure*}[h]
    \centering
    \includegraphics[width=1.0\linewidth]{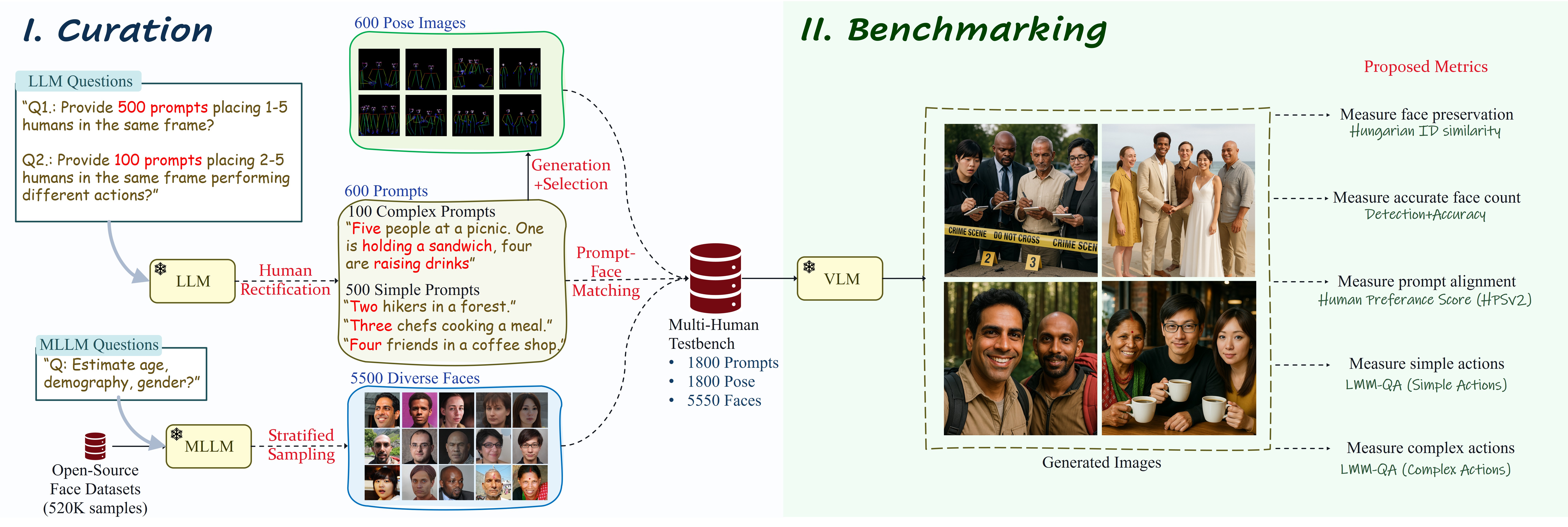}
    %\vspace{- 1.2 em}
    \caption{\textbf{MultiHuman Testbench}. Our MultiHuman Testbench consists of 5,550 IDs across 1,800 samples, including captions describing a scene with of 1-5 humans.}%, 
    \label{fig:main}
    \vspace{-4pt}
\end{figure*}

%\vspace{-5pt}

% Rapid advancements in text-to-image generative models, including subject-driven generation, have revolutionized content creation. They enable users to synthesize photorealistic images from captions and personalize generated content by incorporating subjects from reference images into novel scenes.

While current text-to-image diffusion models can generate high fidelity images, generating scenes featuring multiple humans (from provided reference images) performing text-described actions still remains a challenge. %\textbf{Multi-Human Generation} is useful in creating unified family pictures, hypothetical gatherings, and enabling visual storytelling. 
It requires simultaneously preserving visual characteristics of multiple subjects, accurately rendering their relative positions and interactions, and seamlessly integrating them into the synthesized background. However, current methods~\cite{xiao2024omnigen, he2024uniportrait, xiao2024fastcomposer} frequently exhibit issues such as identity blending, generating the incorrect number of humans, or difficulties in composing the scene according to the text. To make the task easier, some works~\cite{kim2024instantfamily, chen2024training} adopt regional priors as an input to the model, such as human poses, bounding boxes or segmentation masks. While this improves performance, it hinders usability as pose or mask information might not be readily available. %due to the additional input. For instance, models relying on pose need users to "find" images that contain multiple people in the desired pose.

A major challenge in multi-human generation is the lack of a comprehensive and standardized benchmark, along with well defined metrics. Existing benchmarks typically focus on single-subject fidelity~\cite{guo2024pulid, borse2025subzero}, general text-to-image quality~\cite{zhou2024geneval}, or multi-object compositional tasks~\cite{kumari2023multi, he2024dreamstory}. However, none of the currently available open benchmarks address the added complexity of generating multiple distinct humans. %We strongly believe that evaluating models on their ability to handle varying numbers of a diverse set of human subjects, performing actions sampled from a mixture of simple to complex scenarios, is essential to understanding their capabilities and limitations.
To address this issue, we introduce MultiHuman-Testbench, a novel and challenging benchmark. It is built upon a dataset of 1,800 samples, which include carefully crafted text prompts describing scenes with 1 to 5 humans, paired with 5,550 reference human faces, sampled from open-source datasets. We ensure diversity in age, ethnicity, and gender. As many current works rely on regional priors in multi-human scenes~\cite{kim2024instantfamily, chen2024training}, we provide pose conditioning images. Additionally, we propose a multi-faceted evaluation framework designed to capture the nuances of multi-human generation. We propose four complementary metrics: Count Accuracy, Hungarian ID similarity, Human Preference Score, Multimodal LLM (MLLM) question-answering to probe the correctness of simple and complex actions. The proposed testbench has four different tasks: 1) Reference-based Multi-Human Generation in the wild. 2) Reference-based Multi-Human Generation with Regional Priors. 3) ID-Consistent Multi-Human Generation without Reference Images. 4) Text-to-Image Multi-Human Generation. We benchmark current models and identify key areas for improvement. Overall, most methods without regional priors struggle in generating the correct number of people. While proprietery models such as GPT-Image-1 generates plausible images, it lacks preserving facial features and has poor ID retention. We also study biases in current models, in terms of gender, age, status, and ethnicity.

Reference-based Multi-Human Generation in the wild, is the most challenging and least restrictive task in our testbench. We propose new techniques (Sec.~\ref{Proposed Approach: Enhancing Existing Methods}) to adapt current methods for improving their performance this task. Specifically, for unified multi-modal architectures \cite{xiao2024omnigen, OpenAIDocsImageGen}, we propose a method to isolate the reference images to impact only a specific region within the latent space. To match each reference image to regions, we propose an implicit Hungarian matching guided by human segmentation. Our method enhances the ability to maintain individual identities, reducing subject leakage and improving ID similarity. We extend our proposed techniques to two models, OmniGen \cite{xiao2024omnigen} and IR-Diffusion \cite{he2024improving}, resulting in our proposed MH-OmniGen and MH-IR-Diffusion.

In summary, our contributions are:
\begin{itemize}\setlength{\itemsep}{-0.2em}
\item Introduction of a novel benchmark for multi-human ID image generation, featuring diverse subjects, text, and pose conditioning.
\item A comprehensive evaluation suite designed to assess multi-human generation fidelity, including people count accuracy, ID similarity, text-alignment, and MLLM-based assessment.
\item An extensive empirical evaluation and thorough analysis of 30 state-of-the-art zero-shot and training-based generative methods on four different tasks.
\item A novel training-free enhancement for existing multi-human generation methods, utilizing regional isolation and matching for improved identity and compositional control.
\end{itemize}

\vspace{-3pt}
\section{Related Work}
%\textbf{Benchmarking Multi-subject generation:}
\vspace{-5pt}
\textbf{Native Text-to-Image models:} Multiple diffusion based models have been proposed recently \cite{sdxl, rombach2022highresolution, realisticvision15, stabilityai2024stable, borse2025disco}. These models exhibit excellent text-to-image generation ability and can be used as base models for generating multiple humans. 

\textbf{Multi-human Generation with native Text-to-Image models:} To generate ID-consistent or subject driven images with text-to-image models, recent works employ auxiliary models such as IP-Adapter \cite{ye2023ip} or ControlNet \cite{zhao2023uni}. There are also tuning-based approaches which exist such as LoRA \cite{hu2022lora} or MudI \cite{tewel2024training} for this purpose. These methods typically fall short on multi-human generation in the wild, without any regional priors.

\textbf{ID-Consistent Multi-Human Generation without reference images:} Methods such as Consistory~\cite{tewel2024training}, DreamStory~\cite{he2024dreamstory}, IR-Diffusion~\cite{he2024improving} and StoryDiffusion~\cite{zhou2024storydiffusion} have recently gained popularity in ID-consistent multi-subject generation without reference images. These methods generate human faces and use these faces to generate multiple images for tasks like storytelling.

\textbf{Multi-Object Generation:} Recent approaches e.g., MS-Diffusion~\cite{wang2024ms}, MIP-Adapter~\cite{huang2025resolving}, Lambda-Eclipse~\cite{patel2024lambda}, have shown significant performance gain for incorporating multiple objects in the same scene. These can include daily items and in some cases, pets such as dogs and cats. However, they struggle to adapt to multi-human generation as zero-shot ID-preservation is a highly challenging task.

\textbf{Multi-human Generation with native Multi-Modal models:} Unified multimodal models, such as OmniGen \cite{xiao2024omnigen}, Show-O \cite{xie2025showo}, OneDiffusion \cite{le2024one} ACE++ \cite{mao2025ace}, GPT-Image-1 \cite{OpenAIDocsImageGen} and JanusFlow \cite{ma2024janusflow} process the text and vision via same transformer backbone and have shown promises for subject-driven generation. These methods input both the reference images and text prompt in a unified token space, removing the need for additional auxiliary task-specific networks such as IP-Adapter/ControlNet. Omnigen \cite{xiao2024omnigen} was further tuned for ID-preservation. Our evaluations show that among all open-source models, Omnigen produces best results. %For multiple ($>3$) humans, current methods struggle to generate correct number of IDs, facial identities might not be preserved or mixed between subjects.

\textbf{Regional Isolation:} For networks generating images using simply text inputs, recent works such as IR-Diffusion \cite{he2024improving} and InstantFamily \cite{kim2024instantfamily} have proposed methods such as image isolation and repositional attention, which successfully localize multiple humans in the scene by isolating them from each other and mapping them to separate regions in the image latent. These methods have shown great promise in reducing leakage between multiple human identities.

% LoRA~\cite{hu2022lora} and MuDI~\cite{jang2024identity} on SDXL~\cite{sdxl}. For each prompt, we tune a single LoRA for all concepts, using the method described in~\cite{jang2024identity}. Next, we benchmark methods trained for general multi-subject generation, MS-Diffusion~\cite{wang2024ms}, MIP-Adpater~\cite{huang2025resolving} and Lambda-Eclipse~\cite{patel2024lambda}. For multi-human generation we report scores on UniPortrait~\cite{he2024uniportrait}, RectifID~\cite{sun2024rectifid}, Fastcomposer~\cite{xiao2024fastcomposer}, OmniGen~\cite{xiao2024omnigen} and GPT-Image-1~\cite{OpenAIDocsImageGen}. We also benchmark methods which require explicit regional priors such as Regional-Prompting~\cite{chen2024training} with PuLID~\cite{guo2024pulid}, and OMG~\cite{kong2024omg} with InstantID~\cite{wang2024instantid}. Furthermore, we evaluate story-based(reference-free) diffusion models, Consistory~\cite{tewel2024training}, DreamStory~\cite{he2024dreamstory} and IR-Diffusion~\cite{yu2025identity} on our proposed benchmark. 

\vspace{- 0.5 em}
\section{Proposed Approach: Enhancing Existing Methods} \label{Proposed Approach: Enhancing Existing Methods}
\vspace{- 0.5 em}

Reference-based Multi-human generation in the wild (Task 1 in \ref{Results}) is a highly challenging problem. It requires to preserve input identity while rendering the complete scene with the correct number of humans performing a described action. Using insights from benchmarking current approaches in Section~\ref{sec:benchmarking}, we observe several limitations, including identity blending or missing identities. To tackle these issues, we propose two techniques: \textcolor{OliveGreen}{Unified Regional Isolation} and \textcolor{OliveGreen}{Implicit Regional Assignment}, that can be flexibly incorporated with existing methods to enhance their quality.

\begin{figure*}[t]
    \centering
    \includegraphics[width=0.95\linewidth]{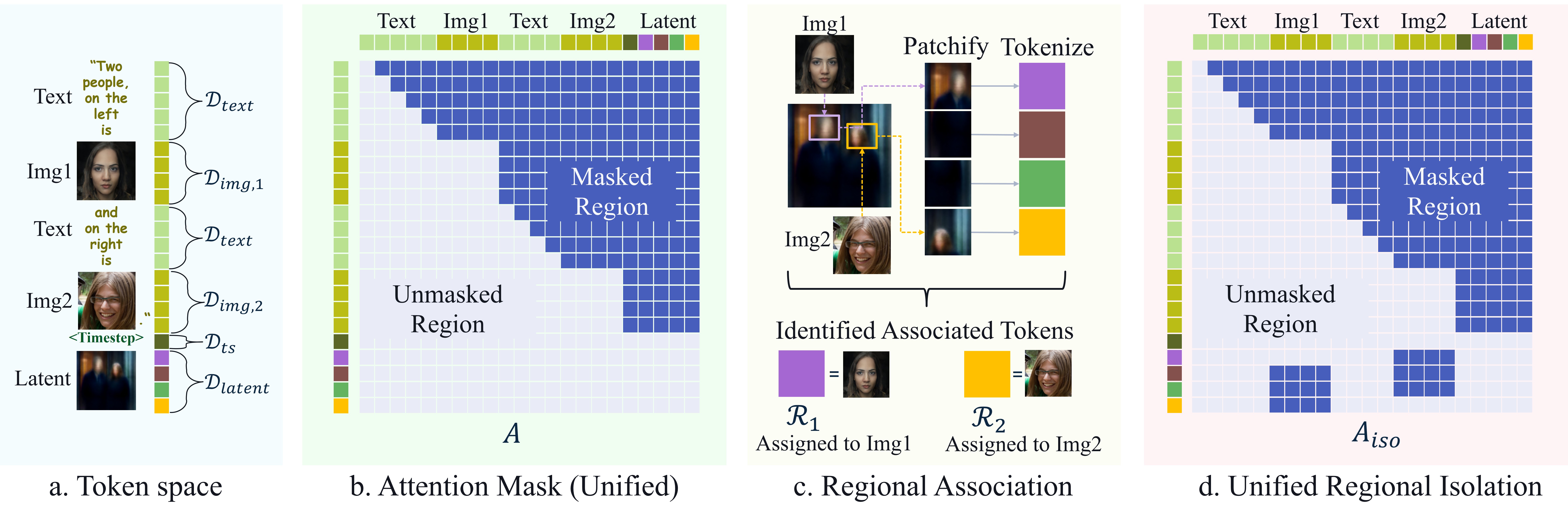}
    %\vspace{- 1.2 em}
    \caption{\textbf{Regional Isolation for Unified Architectures}. The updates to the attention mask for regional isolation are illustrated in the differences between Fig.b and Fig.d.}%, 
    \label{fig:regional_isolation}
    \vspace{- 0.5 em}
\end{figure*}

% \subsection{Tuning-free Multi-Subject generation with Text-to-Image Architectures}
% To generate ID-consistent or subject driven images with models which natively input text, recent works employ auxiliary models such as IP-Adapter or ControlNet. 

\textbf{Unified Regional Isolation:} 
%Due to the promising results of GPT-Image-1 and OmniGen models on multi-human generation, we devise a training-free regional and image isolation-based attention mechanism for unified attention models.
Motivated by~\cite{kim2024instantfamily, he2024improving} for T2I architectures, we develop a regional isolation masking strategy to tackle the limitations relating to identity blending and missing, tailored for unifed models such as OmniGen~\cite{xiao2024omnigen}.

Consider the token space for a unified multimodal model represented in Figure~\ref{fig:regional_isolation} (a). Let $L$ be the total sequence length, and let $i, j \in \{1, \dots, L\}$ be the indices for query and key/value tokens. We define disjoint sets of indices for each token type: $\mathcal{D}_{\text{text}}$, $\mathcal{D}_{\text{img}}$, $\mathcal{D}_{\text{ts}}$(for timestep), and $\mathcal{D}_{\text{latent}}$. Using the setup from OmniGen~\cite{xiao2024omnigen} and Show-o~\cite{xie2025showo}, the self-attention mask $\mathbf{A}$ ($L \times L$) is constructed based on the type of the query token $i$: causal attention for text queries, and bidirectional attention for non-text queries (image, timestep, latent). This is represented as:
\vspace{-4pt}
\[
A_{ij} = \begin{cases}
1 & \text{if } i \in \mathcal{D}_{\text{text}} \text{ and } j \le i \quad (\text{text query: causal}) \\
1 & \text{if } i \notin \mathcal{D}_{\text{text}} \quad (\text{non-text query: bidirectional}) \\
0 & \text{otherwise}
\end{cases}
\]

Consider the tokens in $\mathcal{D}_{\text{img}}$ are derived from $N$ distinct original input images, $\{I_1, \dots, I_N\}$. For each image $I_k$ ($k=1, \dots, N$), let $\mathcal{D}_{\text{img}, k} \subseteq \mathcal{D}_{\text{img}}$ be the set of sequence indices corresponding to its derived tokens. These sets partition $\mathcal{D}_{\text{img}}$. This is represented in Figure~\ref{fig:regional_isolation} for ($k=1, 2$).
For each reference image $I_k$, consider that we find a region of interest (ROI) set $\mathcal{R}_k \subseteq \mathcal{D}_{\text{latent}}$. Now, we construct a new attention mask $\mathbf{A_{iso}}$ ($L \times L$) such that it isolates the images $I_k$ to only the specific region $\mathcal{R}_k$ within the latent. Hence, our proposed attention mask is computed as:
\vspace{-4pt}
\[
A_{iso, ij} = \begin{cases}
1 & \text{if } i \in \mathcal{D}_{\text{text}} \text{ and } j \le i \quad (\text{text query: causal}) \\
1 & \text{if } i \in \mathcal{D}_{\text{img}} \text{ and } (j \notin \mathcal{D}_{\text{latent}} \text{ or } j \in \mathcal{R}_k \text{ where } i \in \mathcal{D}_{\text{img}, k}) \quad (\text{image query: ROI attention}) \\
1 & \text{if } i \in \mathcal{D}_{\text{ts}} \cup \mathcal{D}_{\text{latent}} \quad (\text{timestep/latent query: bidirectional}) \\
0 & \text{otherwise.}
\end{cases}
\]
\textbf{Implicit Region Assignment:}
To construct the attention mask $\mathbf{A_{iso}}$, we need region of interest for every image $\mathcal{R}_k$. This can be done explicitly as in recent methods InstantFamily~\cite{kim2024instantfamily} and Regional Prompting~\cite{chen2024training}, or using a regional prior (pose conditioning or bounding boxes). However, this severely hinders usability, as the users might not want to seek for a multi-human pose image resembling the one which they wish to generate. Hence, to facilitate the generation of multi-human images in the wild, we propose an implicit region assignment strategy that utilize intermediate attention scores and Hungarian matching to assign each reference image to a selected region-of-interest. 

Below, we discuss adaptation of our proposed techniques for different models including Omingen \cite{xiao2024omnigen} and IR-Diffusion \cite{he2024improving}. See Appendix~\ref{sec:appendixmethod} for further details and algorithms.

\textbf{\textcolor{OliveGreen}{MH-Omnigen}:} To find optimal regions for  architectures such as Omnigen \cite{xiao2024omnigen} which have a unified token space for text and reference images, we probe the backbone transformer model at an intermediate timestep. The self-attention maps in the backbone transformer model provide information for the regional overlap for reference images, and the segmentation masks of the intermediate latents provide regional information for each generated person. We perform hungarian matching to find reference inputs with the maximum self-attention region, to eventually find $\mathcal{R}_k$.

\textbf{\textcolor{OliveGreen}{MH-IR-Diffusion}:} In the case for IR-Diffusion~\cite{he2024improving}, the region-of-interest is defined by the models initially generated images. Similar to the original work, we use a segmentation model, SAM2~\cite{ravi2024sam} to generate the region proposals for generated faces. Next, we compute Arcface similarity between generated faces and reference faces to find the best match, and utilize hungarian matching to assign segments of the matched faces as regions $\mathcal{R}_k$.

\begin{wrapfigure}{r}{0.5\linewidth} % r = right side, 0.4\linewidth = width of the figure
    \centering % Center the image within the allocated wrapfigure space
    \includegraphics[width=\linewidth]{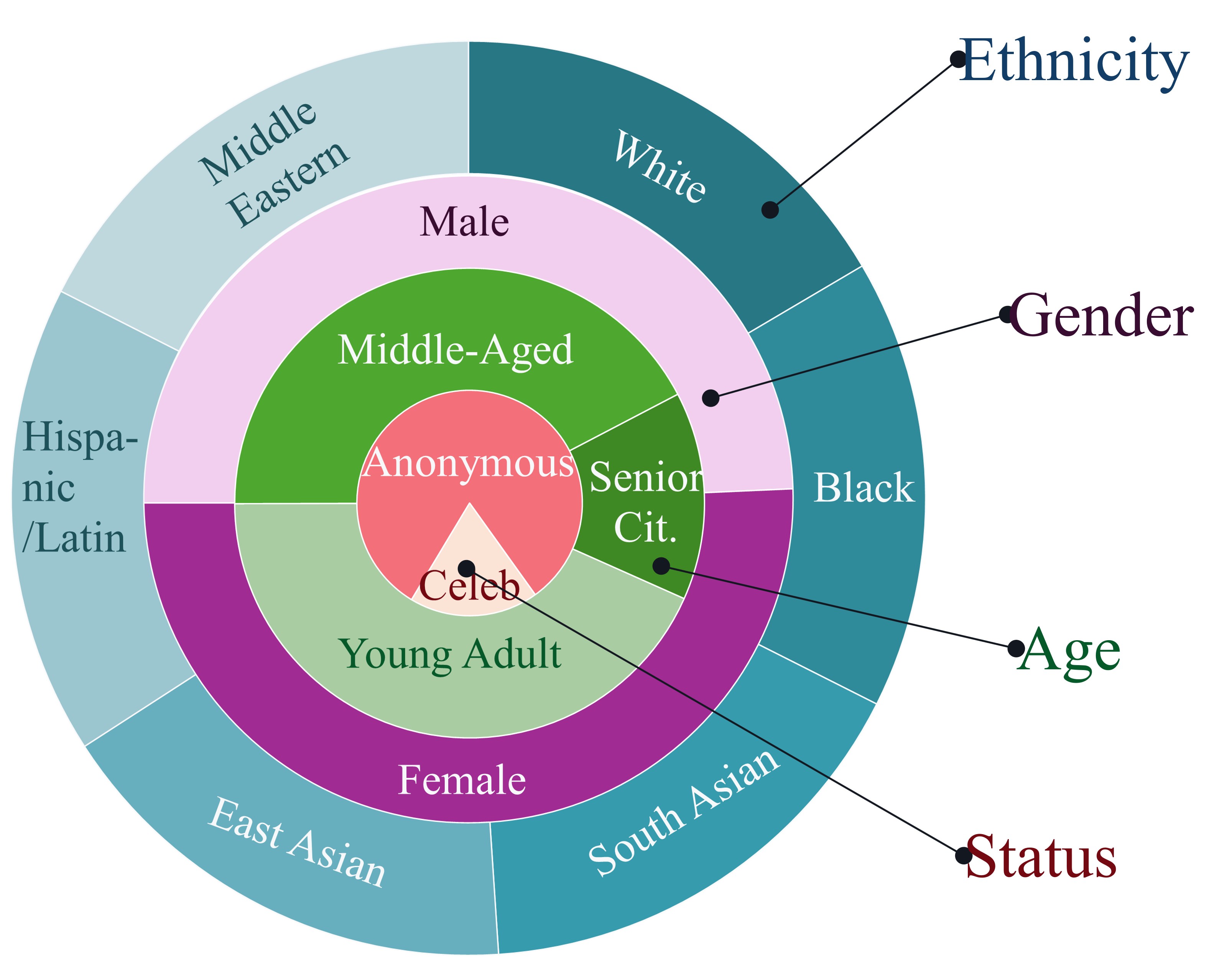} % Image takes the full width of the wrapfigure box
    \caption{\textbf{Data distribution} among four major attributes: Ethnicity, Age, and Gender, Status. See Appendix~\ref{sec:appendixdata} for details.}%,
    \label{fig:distribution}
    \vspace{-30pt}
\end{wrapfigure}

 Our experiments in Sec.~\ref{sec:benchmarking} show that Regional Isolation and Implicit Assignment are training-free plug-and-play methods which can effectively improve different baselines. Due to the implicit matching of identities and localization, we get improved ID similarity with reduced subject blending artifacts.

\section{MultiHuman Testbench}

Below we elaborate the process of curation of our proposed testbench, and discuss different metrics. 
% In this Section, we describe in detail the process of curating and sampling our dataset. We first perform Image selection, followed by Prompt Curation and Assignment. Next, we use a method of generation and human selection to curate pose images for every prompt. Lastly, we perform multi-view generation to provide multiple views of the same face for training-based approaches.

\subsection{Image Selection}
\vspace{-6pt}
We curate images using three existing large-scale image datasets, FFHQ~\cite{karras2019stylebased}, SFHQ~\cite{abdal2020styleflow} and CelebaHQ~\cite{karras2018progressive}, which initially contained approximately 520k samples. These datasets underwent a multi-stage filtering process, where initially we removed human IDs deemed non-"identifiable", using MLLM~\cite{li2024llava} VQA. We prompt the MLLM with the question {\sffamily \textcolor{MyDarkOlive}{"Is the person's face identifiable AND unobstructed?"}} The images with negative response are filtered out. Subsequently, we identify multi-face images using face detection and eliminate them. These steps reduced the dataset size to 94k distinct human face images. For annotation, we employed Gemini Flash 2.0~\cite{gemini-flash-2025} to classify each of these images based on three predefined attributes: estimated age-bracket (one of 16-35, 35-60, 60+), estimated demographic (sorted into 6 categories: Caucasian/White, Black/Native African, South-Asian, East-Asian, Hispanic/Latin, Middle Eastern/North African), and Estimated Gender assigned at birth (divided into 2 categories: Male, Female). Our dataset does not contain any images of minor subjects. Following this annotation, a non-biased test set consisting of 5,550 images was generated from the curated and labeled dataset using a stratified sampling approach to ensure representative distribution across the various age, ethnicity, and gender buckets. See Figure~\ref{fig:distribution} for the final distribution. For ethnicity and gender, we provided a target uniform distribution. For age-bracket, we set a target percentage of 15\%, using insights from PopulationPyramid~\cite{populationpyramid2024}.

\subsection{Prompt Curation and Assignment}
\begin{figure*}[t]
    \centering
    \includegraphics[width=1.0\linewidth]{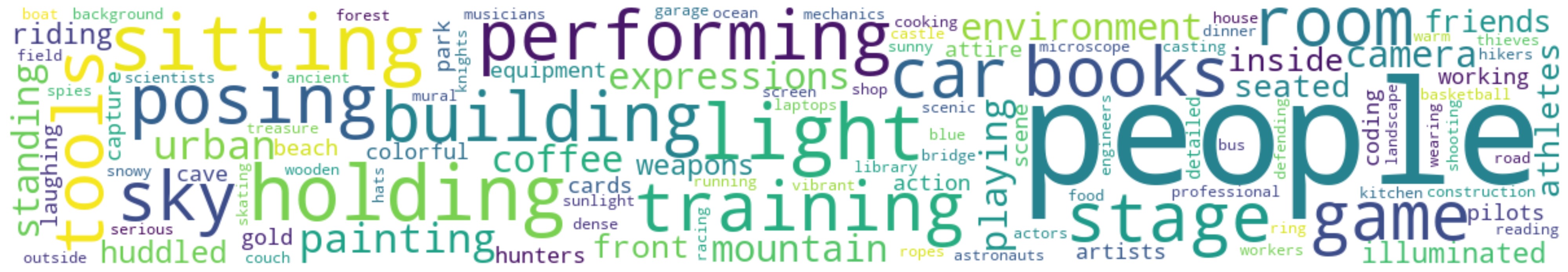}
    %\vspace{- 1.2 em}
    \caption{\textbf{Wordcloud}. The graphic shows words from our caption space.}%, 
    \label{fig:wordcloud}
    \vspace{- 1 em}
\end{figure*}

\vspace{-6pt}
We aim to create a diverse set of prompts and begin by using Gemini Flash 2.0~\cite{gemini-flash-2025} to generate an initial set of 100 prompts, each describing five people performing the same action. This set was then recontextualized to create variations for scenarios involving one to four people, forming the core of our "simple prompt" set. In parallel, we curated a separate collection of 25 prompts specifically designed to depict multiple people performing different actions within the same image, which were also recontextualized for scenarios involving two to four people, resulting in our "complex prompt" set. Our prompt set comprises the combination of these two collections, totaling 125 distinct prompts. The simple prompts cover scenarios with one to five people, while the complex prompts are intended for two to five people. To build MultiHuman Testbench, we generated testing samples by sampling three random iterations of human IDs for each prompt, resulting in a total of 1,800 unique testing samples. Each prompt underwent careful human revision after multiple iterations of generation with various subject-driven generation models to ensure the final set's quality and suitability. The wordcloud for our final prompt action space is visible in Figure~\ref{fig:wordcloud}. To populate the 1,800 testing samples, each requiring a specific human ID from the 5,550-image test set, a distributed sampling approach was employed to assign the 5,550 human IDs to the prompt iterations.

\subsection{Pose Image Estimation}
\vspace{-6pt}
Pose images act as Regional Priors for benchmarking in Task3 (see Sec.~\ref{Results}). Hence, to obtain suitable pose information for the MultiHuman Testbench, we generate pose from two sources: a) the best results for each prompt using our outputs from Tasks 1 and 4, and b) a Text-to-Pose generation~\cite{bonnet2024text} model. Using resulting Images from Tasks 1 and 4, we find all images with the best Count and Action metrics for each prompt, and filter the ones which fall below a specific threshold (i.e. below 0.97 for action similarity, and below 1 for Count Accuracy). For prompts which ar filtered out due to poor overall scores, we obtain pose information using Text-to-pose~\cite{bonnet2024text}. We generate 20 distinct pose samples for each prompt. A crucial human selection step was then performed, where reviewers carefully evaluated each generated option to identify the single best pose for every prompt in our dataset. This rigorous manual filtering was necessitated by the inherent limitations of the text-to-pose generation model (along with limitations of Task 1 and Task 4), which occasionally demonstrate susceptibility to generating erroneous poses. The results presented in Appendix~\ref{sec:appendixquant} provide strong evidence for the effectiveness of our provided pose pairs, yielding significantly improved results across most metrics.

\subsection{Multi-View Image Generation}
\vspace{-6pt}
For each human ID in the curated dataset, we utilized the PuLID-Flux~\cite{guo2024pulid} to generate a collection of five distinct images. We prompt the model to render the person's identity in diverse contexts, capturing various perspectives such as full-portrait and side views, and placing them in differing environments. The intent behind these multiple generated images per ID is to gather comprehensive training data to improve performance on tuning-based multi-human generation models.

\subsection{Metrics}
\vspace{-6pt}
To evaluate multi-human image generation, our benchmark proposes a suite of metrics specifically designed to capture various critical aspects of the generated output. 

\textbf{Hungarian ID Similarity.} We propose an ID similarity metric using ArcFace embeddings \cite{deng2019arcface}. To match input and generated IDs in the multi-human setting, we use cosine similarity of Arcface embeddings, and use the hungarian algorithm~\cite{Kuhn1955} to match each face while maximizing cost. The Hungarian ID similarity for a given image is thus the average matched ID similarity.

% \begin{algorithm}
% \caption{Calculate Identity Similarity Metric ($S_{id}$)}
% \label{alg:identity_metric}
% \begin{algorithmic}[1] % The [1] adds line numbering
% \Require A set of $N$ reference face embeddings: $\mathbf{F}^{\text{ref}} = \{\mathbf{f}_i^{\text{ref}} \mid i=1, \dots, N\}$
% \Require A set of $M$ generated face embeddings: $\mathbf{F}^{\text{gen}} = \{\mathbf{f}_j^{\text{gen}} \mid j=1, \dots, M\}$
% \Ensure The average identity similarity metric: $S_{id}$

% \State Initialize an $N \times M$ similarity matrix $\mathbf{S}$.
% \For {$i = 1, \dots, N$}
%     \For {$j = 1, \dots, M$}
%         \State Calculate cosine similarity $s_{ij} = \cosSim{\mathbf{f}_i^{\text{ref}}}{\mathbf{f}_j^{\text{gen}}}$. % Referencing the formula defined earlier
%         \State Set $\mathbf{S}_{ij} = s_{ij}$.
%     \EndFor
% \EndFor
% \State
% \State Formulate an $N \times M$ cost matrix $\mathbf{C}$ where $c_{ij} = -s_{ij}$ for all $i, j$.
% \State
% \State Apply the Hungarian algorithm to the cost matrix $\mathbf{C}$ to find a binary assignment matrix $\mathbf{X}$ ($N \times M$).
% \State
% \State Calculate the average identity similarity $S_{id}$:
% \State $S_{id} = \frac{1}{N} \sum_{i=1}^N \sum_{j=1}^M X_{ij} s_{ij}$ % Referencing the formula defined earlier
% \State
% \State \Return $S_{id}$

% \end{algorithmic}
% \end{algorithm}

Consider a set of $N$ input face images, indexed by $i=1, \dots, N$, and a set of $M$ output face detections in the generated image, indexed by $j=1, \dots, M$. Consider Arcface embeddings for input images $\mathbf{F}^{\text{ref}} = \{\mathbf{f}_i^{\text{ref}} \mid i=1, \dots, N\}$, where $\mathbf{f}_i^{\text{ref}} \in \mathbb{R}^d$ and for generated faces $\mathbf{F}^{\text{gen}} = \{\mathbf{f}_j^{\text{gen}} \mid j=1, \dots, M\}$, where $\mathbf{f}_j^{\text{gen}} \in \mathbb{R}^d$. Here, $d$ is the dimensionality of the feature space. Next, we define the similarity $s_{ij}$ between reference face $i$ and generated face $j$ using cosine similarity:
\vspace{-4pt}
\[ s_{ij} = cosSim(\mathbf{f}_i^{\text{ref}}, \mathbf{f}_j^{\text{gen}}) = \frac{(\mathbf{f}_i^{\text{ref}})^\top \mathbf{f}_j^{\text{gen}}}{\| \mathbf{f}_i^{\text{ref}} \|_2 \| \mathbf{f}_j^{\text{gen}} \|_2} \]

\vspace{-10pt}

We form an $N \times M$ similarity matrix $\mathbf{S}$, where $\mathbf{S}_{ij} = s_{ij}$. Since the Hungarian algorithm finds a minimum cost assignment, we define the cost $c_{ij}$ as the negative similarity, $c_{ij} = -s_{ij}$. Using the Hungarian algorithm, we find a binary assignment matrix $\mathbf{X}$ ($X_{ij}=1$ if matched, $0$ otherwise).
% \begin{align*}
% \min_{\mathbf{X}} &\quad \sum_{i=1}^N \sum_{j=1}^M c_{ij} X_{ij} \\
% \text{subject to} &\quad \sum_{j=1}^M X_{ij} \le 1 \quad \text{for } i = 1, \dots, N \quad (\text{each reference matched to at most one generated}) \\
% &\quad \sum_{i=1}^N X_{ij} \le 1 \quad \text{for } j = 1, \dots, M \quad (\text{each generated matched to at most one reference}) \\
% &\quad X_{ij} \in \{0, 1\} \quad \text{for } i = 1, \dots, N, j = 1, \dots, M
% \end{align*}
For each reference input $i$, if reference $i$ is matched to a generated face $j$ (i.e., $\sum_{k=1}^M X_{ik}=1$), its contribution to the ID metric is the similarity $s_{ij}/N$ for the matched $j$. If reference $i$ is not matched to any generated face (i.e., $\sum_{k=1}^M X_{ik}=0$), its contribution to the ID metric is $0$. Hence, the average similarity over all $N$ reference inputs, denoted $S_{id}$, is denoted as: $S_{id} = \frac{1}{N} \sum_{i=1}^N \sum_{j=1}^M X_{ij} s_{ij}$.
Our proposed Hungarian ID similarity metric objectively evaluates the model's effectiveness in maintaining consistent and uniquely recognizable identities across different generations. Further, the proposed metric penalizes for subject/ID mixing.

\textbf{Count Accuracy.} Next, we assess the accuracy of the generated people count. This verifies the model's ability to precisely adhere to the numerical specification in the prompt. We use a face detection model \cite{deng2020retinaface} to count detected human faces in the generated image. Hence, as $N$ is the number of reference images and $M$ is the number of generated faces, the count accuracy is $S_{\text{count}} = \delta_{MN}$, where $\delta$ is the kronecker delta~\cite{garfken67:math} function.

\textbf{Quality/Prompt Alignment.} Third, text alignment for overall scene consistency is evaluated using the HPSv2 score \cite{wu2023human}, $S_{\text{hps}}$. This metric goes beyond individual elements to measure how well the entire generated image corresponds to the textual description of the scene, ensuring that contexts, environments, and overall narrative specified in the prompt are accurately reflected. 

\textbf{MLLM Action QA.} Fourth, to probe the correctness of simple and complex actions and interactions among multiple individuals, we utilize Multimodal Large Language Model (MLLM) question-answering. This approach allows for a deeper semantic evaluation by querying the MLLM about specific details, activities, and relationships depicted in the generated image, thereby assessing challenging compositional aspects. We propose to report the average separately for simple actions (Action-S) and complex actions (Action-C), as they provide deeper meaning. To generate the questions, we probe Gemini-Flash~\cite{gemini-flash-2025} to extract actions from each text prompt, and re-contextualize these into questions. For instance, assume the prompt is {\sffamily \textcolor{MyDarkOlive}{"Five people caroling during winter: among them, two people are holding song books, and three people are singing"}}. For this prompt, the questions which are generated to rank complex actions are as follows: {\sffamily \textcolor{MyDarkOlive}{"Q1: Are two people in this image holding song books? Choices: 1(No), 10(Yes), 5(Partially)? Q2: Are the people in this image caroling? Choices: 1(No), 10(Yes), 5(Partially)?"}}. Hence, the final (Action-C) score is the average score.

%\textbf{LMM Quality QA.} Finally, we employ an LMM for providing an assessment of the overall generated quality. This metric extends beyond simple visual fidelity or similarity, allowing the LMM to provide a more holistic evaluation that considers factors like naturalness, plausibility, aesthetic quality, and the successful integration of all elements from a human-like perspective. Collectively, these five metrics form a comprehensive evaluation framework, enabling a detailed and multifaceted analysis of multi-human image generation performance.

\textbf{Unified Evaluation Metric.} While individual metrics provide granular insights, practical model comparison benefits from a unified score. We propose a composite metric that integrates identity fidelity and prompt-image alignment. Specifically, we compute the geometric mean of (i) Hungarian ID similarity $S_{\text{id}}$ and (ii) a weighted aggregate of alignment metrics (HPS, Action-S, Action-C, and Count Accuracy). The geometric mean ensures that poor performance in either dimension significantly reduces the overall score, reflecting the interdependence of identity preservation and semantic correctness.

Let:
\[
S_{\text{align}} = \frac{S_{\text{hps}} + S_{\text{act}} + S_{\text{count}}}{3}.
\]
To emphasize alignment, we apply a quadratic weighting to $S_{\text{align}}$ within the geometric mean. The unified metric $S_{\text{U}}$ is:
\[
S_{\text{U}} = \big( S_{\text{id}} \times (S_{\text{align}})^2 \big)^{\frac{1}{3}}.
\]
This formulation ensures that if either identity preservation or alignment is weak, the overall score remains low, promoting balanced optimization across all aspects of multi-human image generation.

\vspace{-6pt}
\section{Benchmarking}
\label{sec:benchmarking}

\begin{figure*}[t]
    \centering
    \includegraphics[width=0.85\linewidth]{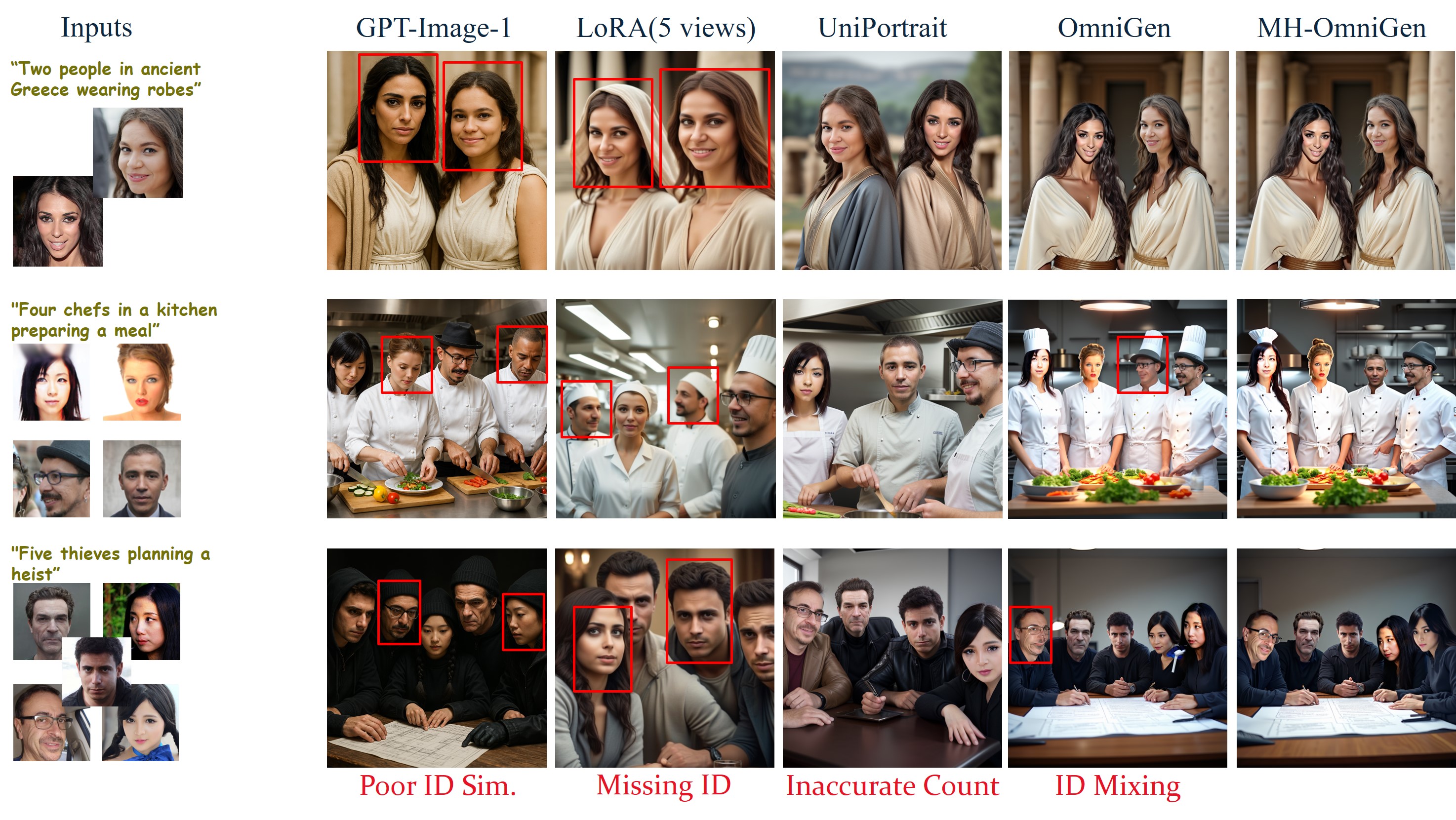}
    \vspace{- 1.2 em}
    \caption{\textbf{Qualitative Results on Multi-Human Generation in the wild}. The image shows the best performing methods: UniPortrait, LoRA, GPT-Image-1, OmniGen and MH-OmniGen.}%, 
    \label{fig:qualitative}
    \vspace{- 0.5 em}
\end{figure*}

\vspace{-6pt}
%\subsection{Experimental Setup and Models}
%\vspace{-6pt}
We benchmark models under four settings: 1). Reference-based Multi-Human Generation in the wild, 2). Reference-based Multi-Human Generation with Regional Priors, 3). Story Consistent Multi-Human generation without reference images and 4). Text-to-Image multi-human generation without reference images. We use tuning-based methods such as LoRA~\cite{hu2022lora} and MuDI~\cite{jang2024identity} on SDXL~\cite{sdxl}. For each prompt, we tune a single LoRA for all concepts, using the method in~\cite{jang2024identity}. Next, we benchmark methods trained for general multi-subject generation, MS-Diffusion~\cite{wang2024ms}, MIP-Adapter~\cite{huang2025resolving} and Lambda-Eclipse~\cite{patel2024lambda}. For multi-human generation, we evaluate UniPortrait~\cite{he2024uniportrait}, RectifID~\cite{sun2024rectifid}, Fastcomposer~\cite{xiao2024fastcomposer}, OmniGen~\cite{xiao2024omnigen}, Flux-Kontext~\cite{labs2025flux1kontext} and GPT-Image-1~\cite{OpenAIDocsImageGen}. We also benchmark methods which require explicit regional priors such as Regional-Prompting~\cite{chen2024training} with PuLID~\cite{guo2024pulid}, and OMG~\cite{kong2024omg} with InstantID~\cite{wang2024instantid}. Furthermore, we evaluate story-based(reference-free) diffusion models, Consistory~\cite{tewel2024training}, DreamStory~\cite{he2024dreamstory} and IR-Diffusion~\cite{yu2025identity} on our proposed benchmark. Finally, we evaluate native text-to-image models, SD-1.5~\cite{rombach2022highresolution}, RV-1.5~\cite{realisticvision15}, SDXL~\cite{sdxl}, SD3.5~\cite{stabilityai2024stable}, Flux~\cite{labs2025flux1kontext} and OmniGen~\cite{xiao2024omnigen}, for generating accurate number of humans. All implementation details and hyperparameters are provided in Appendix~\ref{sec:appendiximpl}.

\subsection{Results}
\label{Results}

\textbf{Task 1. Reference-based Multi-Human Generation in the Wild:} 

Our results for task 1 are summarized in Table~\ref{tab:multihuman_ref}. We evaluate the performance of four different types of models: Proprietary, Tuning-based, Multi-Object Tuning-Free and Multi-Human Tuning-Free methods. From the scores, we find that the performance of each method is significantly influenced by the backbone model it builds upon.  On average, \textbf{Multi-object Tuning-Free methods} perform worse compared to other approaches. This is because generating humans and keeping their likeness intact is a significantly challenging problem, compared to the objects the methods have been trained on. Next, \textbf{Tuning-based methods} MuDI~\cite{jang2024identity} and LoRA~\cite{hu2022lora} perform slightly better, but are significantly bounded by the base architecture SDXL. As observed, training with 5 views generated from PuLID~\cite{guo2024pulid} performs better than a single view. Moving to \textbf{Multi-Human Tuning-free} approaches, we can observe that UniPortrait~\cite{he2024uniportrait}, built on RV1.5, significantly outperforms other SD1.5-based approaches such as RectifID~\cite{sun2024rectifid} and FastComposer~\cite{xiao2024fastcomposer} across all metrics. However, unified multi-modal models, OmniGen~\cite{xiao2024omnigen} and GPT-Image-1~\cite{OpenAIDocsImageGen}, perform significantly better than the rest. Notably, our proposed MH-OmniGen consistently outperforms its predecessor across four of the five metrics. We observe a \textbf{5.1} point difference in Multi-ID and \textbf{4.1} point difference in action similarity. This validates the effectiveness of our Unified Regional Isolation and Implicit Assignment method. Finally, among all evaluated methods, GPT-4o (via GPT-Image-1) achieves the highest overall performance in count accuracy, HPS, and action-based metrics. However, its performance in ID similarity is notably weaker (\textbf{25.7} points) than MH-OmniGen. This is due to the fact that GPT hallucinates features on humans, and in many cases isn't able to effectively maintain the identity of the person. When considering the Unified metric, which balances all aspects of performance, MH-OmniGen achieves the best score (\textbf{61.6}), followed by OmniGen (\textbf{59.2}) and GPT-Image-1 (54.3), demonstrating that our proposed method achieves the strongest overall balanced performance across all dimensions. \textbf{Overall}, we want to stress that \textbf{None} of the methods perform consistently well at a high standard for this task in terms of visual quality. Within the open-source methods, \textbf{None} of the models can consistently generate images with a high Action-C or Count score. There is significant scope for improvement in this setting.

\textbf{Qualitative Results:} In Figure~\ref{fig:qualitative}, we show visual Results for the best models performing Task 1. As observed in this image, GPT-1 isn't able to effectively maintain human ID, owing to poor scores. On the other hand, Uniportrait generates good results but often with the inaccurate number of humans. OmniGen results have artifacts related to ID mixing, which are considerably repaired in MH-OmniGen. However, OmniGen-based methods tend to "copy" human faces.  It is important to note that these are some of the better looking images for each method. We share more visual results in the Appendix, highlighting a heavy scope for improvement.

\begin{table}[htbp]
    \renewcommand{\arraystretch}{1.5}
    \fontsize{8.0pt}{6.75pt}\selectfont
    \centering
    \begin{tabular}{l|ll| ccccc c} % Added one more column for Unified metric
        \hline
         & \multirow{2}{*}{Backbone} & \multirow{2}{*}{Model} & \multicolumn{6}{c}{Metrics} \\ % Updated colspan to 6
        & & & Count & Multi-ID & HPS & Action-S & Action-C & Unified \\ % Added Unified here
        \hline
        \multicolumn{9}{c}{\cellcolor{lightyellow}\textbf{Task 1}: Reference-based Multi-Human Generation in the Wild} \\
        \rowcolor{lightpurple} Proprietary & GPT-4o & GPT-Image-1 & 87.9 & 28.8 & 30.3 & 97.0 & 91.1 & 54.3 \\
        \cline{2-9}

        \hline
        \cellcolor{lightblue} & \multirow{4}{*}{SDXL} & LoRA(1 view) & 47.3 & 20.2 & 25.3 & 61.0 & 55.4 & 36.2 \\
        \cellcolor{lightblue} & & LoRA(5 views) & 52.6 & 22.0 & 25.9 & 73.0 & 72.9 &  41.0 \\
        \cellcolor{lightblue} & & MuDI(1 view) & 48.1 & 23.6 & 24.8 & 64.0 & 51.5 & 37.7 \\
        \cellcolor{lightblue} \multirow{-4}{*}{\parbox{1.7cm}{Tuning-Based}} & & MuDI(5 views) & 53.9 & 24.6 & 25.6 & 67.3 & 71.5 & 42.3 \\
        \hline
        \cellcolor{lightred} & & IP-Adapter & 34.3 & 9.3 & 23.2 & 49.6 & 46.9 & 16.3 \\
        \cellcolor{lightred} & \multirow{-2}{*}{SDXL} & MIP-Adapter & 39.2 & 11.9 & 24.0 & 57.6 & 53.7 & 19.2 \\
        \cdashline{2-9}
        \cellcolor{lightred} \multirow{-3}{*}{\parbox{1.7cm}{\centering Multi-Object\\ Tuning-Free}} & Kand2.2 & Lamda-Eclipse & 53.3 & 12.5 & 23.4 & 56.1 & 50.8 & 23.1 \\

        \hline
        \cellcolor{lightgreen} & RV1.5 & UniPortrait & 58.5 & 44.2 & 25.9 & 76.2 & 67.2 & 51.7 \\
        \cdashline{2-9}
        \cellcolor{lightgreen} & \multirow{2}{*}{SD1.5} & RectifID & 37.8 & 18.6 & 24.8 & 67.3 & 68.2 & 33.8 \\
        \cellcolor{lightgreen} & & Fastcomposer & 31.2 & 12.2 & 21.7 & 48.9 & 41.2 & 20.2 \\
        \cdashline{2-9}
        \cellcolor{lightgreen} & \multirow{2}{*}{Phi-3} & OmniGen & \textbf{60.5} & \underline{49.4} & \underline{26.2} & \underline{87.5} & \underline{71.3} & \underline{59.2} \\
        \cellcolor{lightgreen} \multirow{-5}{*}{\parbox{1.7cm}{\centering Multi-Human\\ Tuning-Free}} & & \cellcolor{Gray}\textcolor{OliveGreen}{MH-Omnigen} & \cellcolor{Gray}\underline{60.3} & \cellcolor{Gray} \textbf{54.5} & \cellcolor{Gray} \textbf{26.3} & \cellcolor{Gray} \textbf{91.6} & \cellcolor{Gray} \textbf{72.9} & \cellcolor{Gray} \textbf{61.6} \\
        \bottomrule
    \end{tabular}
    \caption{Multi-Human Generation with Reference Images in the wild.}
    \label{tab:multihuman_ref}
\vspace{-5pt}
\end{table}

\textbf{Task 2. Reference-based Multi-Human Generation with Regional Priors:}

Next, we evaluate methods for Task 2, which focuses on reference-based multi-human generation with regional priors. Table~\ref{tab:multihuman_ref_pose} shows results for Tuning-based, Multi-Object Tuning-Free, and Multi-Human Tuning-Free methods, all leveraging pose or box priors. As observed, the introduction of our provided human-rectified pose priors significantly improves quantitative metrics, particularly Count accuracy, compared to Task 1. We also observe a general increase in Action scores. For instance, MIP-Adapter~\cite{huang2025resolving} shows significantly higher Count accuracy with pose priors. Within the Multi-Human Tuning-Free group, we observe varied performance across metrics depending on the backbone and specific prior usage. RectifID~\cite{sun2024rectifid} achieves the highest Count accuracy (\textbf{90.1}), while OMG-InstantID~\cite{kong2024omg} (SDXL) excels in HPS (\textbf{27.2}) and Action scores (\textbf{90.4}, \textbf{78.9}), and Regional-PuLID~\cite{chen2024training} (Flux) shows the strongest Multi-ID retention (\textbf{50.7}). Flux-Kontext, also based on the Flux backbone, demonstrates strong performance in Action scores (80.9, \textbf{79.8}), achieving the highest Action-C score among all methods, while maintaining competitive Count accuracy (76.8) and HPS (26.9). OmniGen demonstrates competitive performance in Task 2, maintaining strong HPS (\textbf{27.4}) and Action scores (\textbf{86.2}) and decent Multi-ID (\textbf{48.2}) when incorporating pose priors. Overall, Task 2 results highlight the significant benefit of regional guidance for key metrics like count, while demonstrating that achieving high performance across all aspects (like ID fidelity, action, and overall quality) remains a challenge. This is due to the fact that different methods are strong in different areas. When considering the Unified metric, which balances all aspects of performance, UniPortrait and OmniGen achieve the highest scores (\textbf{62.5}), followed by Regional-PuLID (56.4), Flux-Kontext (55.7), and RectifID (55.4), demonstrating that multi-human tuning-free methods generally outperform tuning-based and multi-object approaches in overall balanced performance.

\begin{table}[htbp]
    \renewcommand{\arraystretch}{1.5}
    \fontsize{8.0pt}{6.75pt}\selectfont
    \centering
    \resizebox{\textwidth}{!}{%
    \begin{tabular}{l|ll| c | cccccc} % Added one extra column for Unified
        \hline
         & \multirow{2}{*}{Backbone} & \multirow{2}{*}{Model} & \multirow{2}{*}{\parbox{1.5cm}{Regional Conditioning}} & \multicolumn{6}{c}{Metrics} \\ % Updated colspan to 6
        & & & & Count & Multi-ID & HPS & Action-S & Action-C & Unified \\ % Added Unified
        \hline
        \multicolumn{10}{c}{\cellcolor{lightyellow}\textbf{Task 2}: Reference-based Multi-Human Generation with Regional Priors} \\
        
        \hline
        \cellcolor{lightblue} & \multirow{2}{*}{SDXL} & LoRA(1 view) & & 85.3 & 17.3 & 26.1 & 73.6 & 78.0 & 40.3 \\
        \cellcolor{lightblue} \multirow{-2}{*}{\parbox{1.7cm}{Tuning-based}} & & LoRA(5 views) & \multirow{-2}{*}{Pose} & \underline{89.6} & 21.4 & 26.0 & 77.7 & 73.6 & 44.1 \\
        \hline
        \cellcolor{lightred} \parbox{1.7cm}{Multi-Object Tuning-Free} & SDXL & MIP-Adapter & Pose & 81.5 & 14.1 & 25.0 & 69.2 & 67.2 & 36.9 \\
        \hline
        \cellcolor{lightgreen} & Flux & Regional-PuLID & Boxes & 67.4 & \textbf{50.7} & 26.1 & 74.1 & 68.0 & 56.4 \\
        \cellcolor{lightgreen} & Flux & Flux-Kontext & Boxes & 76.8 & 39.2 & 26.9 & 80.9 & \textbf{79.8} & 55.7 \\
        \cdashline{2-10}
        \cellcolor{lightgreen} & RV1.5 & UniPortrait & Pose & 78.3 & \underline{49.2} & 26.3 & \underline{88.2} & 78.1 & \textbf{62.5} \\
        \cellcolor{lightgreen} & SD1.5 & RectifID & Pose & \textbf{90.1} & 26.4 & 25.7 & 78.7 & 73.5 & 55.4 \\
        \cellcolor{lightgreen} & SDXL & OMG-InstantID & Pose & 71.2 & 32.6 & \underline{27.2} & \textbf{90.4} & \underline{78.9} & 54.6 \\
        \cellcolor{lightgreen} \multirow{-6}{*}{\parbox{1.7cm}{Multi-Human Tuning-Free}} & Phi-3 & OmniGen & Pose & 77.2 & 48.2 & \textbf{27.4} & 86.2 & 75.3 & \textbf{62.5} \\
        \bottomrule
    \end{tabular}}
    \caption{Multi-Human Generation with Reference Images, with regional priors}
    \label{tab:multihuman_ref_pose}
\vspace{-5pt}
\end{table}

\textbf{Task 3. ID-Consistent Multi Human Generation without reference images:}

Table~\ref{tab:multihuman_noref} displays results on Task 3. We benchmarked four approaches. The performance varies across metrics, with ConsiStory and DreamStory showing lower accuracy in Count and ID-Similarity compared to IR-Diffusion and MH-IR-Diffusion. Notably, MH-IR-Diffusion achieved the highest scores in both Count (\textbf{62.6}) and Multi-ID (\textbf{33.3}). Due to the process of masking and hungarian assignment, the model is successfully able to preserve ID information while keeping the remaining metrics stable. While IR-Diffusion led slightly in Action-S (\textbf{86.3}), and all models performed similarly on HPS. We observe for the Complex Action metrics, that the performance reduces slightly as ID similarity improves. This is due to the fact that the results are closer to original model generation, as lesser ID's have been matched. Overall, we see a significant scope of improvement for every method in this list, due to poor performance on all metrics.

\begin{table}[htbp] % [htbp] are placement specifiers (here, top, bottom, page)
    \renewcommand{\arraystretch}{1.5}
    \fontsize{8.0pt}{6.75pt}\selectfont
    \centering % Center the table on the page
    \begin{tabular}{ll|c|ccccc} % Defines column format:
        % l: Left-aligned text column
        % c: Centered column (good for numbers/metrics)
        % Total columns: Model (l) + Backbone (l) + Pose Input (c) + Accuracy (c) + ID Sim (c) + HPS (c) + Quality (c) + Action Rec (c) = 8
        \hline % Top horizontal rule from booktabs
         \multirow{2}{*}{Backbone} & \multirow{2}{*}{Model} &  \multirow{2}{*}{Resolution} & \multicolumn{5}{c}{Metrics}\\
        %\cmidrule(lr){4-6} \cmidrule(lr){7-8} % Horizontal rules below multi-column headers
        % These cmidrules span columns 4-6 and 7-8 with trimming (lr)
        & & & Count & Multi-ID & HPS & Action-S & Action-C \\
        \hline
        \multicolumn{8}{c}{\cellcolor{lightyellow}\textbf{Task 3}: ID-Consistent Multi-Human Generation without Reference Images} \\ 
        
        \hline  % Middle horizontal rule from booktabs
        & ConsiStory & & 44.6 & 16.2 & 28.0 & 84.1 & \textbf{71.9} \\     
        & DreamStory & & 45.0 & 19.7 & 28.2 & 84.8 & 71.8 \\
        & IR-Diffusion & & \underline{62.4} & \underline{27.6} & \underline{29.4} & \textbf{86.3} & \underline{71.8} \\
        \multirow{-4}{*}{Playground-v2.5} & \cellcolor{Gray} \hspace{-0.4em}\textcolor{OliveGreen}{MH-IR-Diffusion} & \cellcolor{Gray} \multirow{-4}{*}{$768 \times 1280$} & \cellcolor{Gray} \textbf{62.6} & \cellcolor{Gray} \textbf{33.3} & \cellcolor{Gray} \textbf{29.2} & \cellcolor{Gray} \underline{85.9} & \cellcolor{Gray} 71.3 \\
           
        % --- End of Empty Rows ---
        \bottomrule % Bottom horizontal rule from booktabs
    \end{tabular}
    \caption{Multi-Human Generation without Reference Images} % Table caption
    \vspace{-5pt}
    \label{tab:multihuman_noref} % Label for cross-referencing
\end{table}

\textbf{Task 4. Text-to-image Multi-Human Generation:} 

For Task 4, we evaluate the overall effectiveness of text-to-image methods on generating humans with accurate count performing text-described simple and complex action. This is mainly to motivate the selection of a suitable base architecture for follow-up methods to build their solutions on. Our results are summarized in Table~\ref{tab:t2i_gen}. Across all methods we evaluated, Flux, SD3.5 and OmniGen perform best, given the fact that they are larger models and have been trained on richer data. Notably, Flux produces images with consistent Count accuracy over 3, 4, 5 humans, compared to the other methods which show a steeper drop in performance after increasing the number of humans. SD3.5 is highly competitive in generating the correct actions (simple and complex), and OmniGen produces the best HPS. Overall, however, there is significant scope for improvement in terms of human count, as the best score (\textbf{63.9}) is still quite low.

\begin{table}[htbp] % [htbp] are placement specifiers (here, top, bottom, page)
    \renewcommand{\arraystretch}{1.5} % Increase vertical space between rows
    \fontsize{8.0pt}{6.75pt}\selectfont % Set font size
    \centering % Center the table on the page
    % Defines column format:
    % l: Left-aligned text column (for Model)
    % |: Vertical rule
    % c: Centered column (for the 7 metrics)
    % Total columns: Model (l) + 7 Metrics (c) = 8
    \begin{tabular}{l|cccc|ccc}
        \hline % Top horizontal rule
        % First header row: Model (spans 2 rows) and Metrics (spans 7 columns)
        \multirow{2}{*}{Model} & \multicolumn{4}{c|}{Person Count} & \multicolumn{3}{c}{Prompt Alignment}\\
        % Second header row: specific metric names under the Metrics header
        % The first column (&) is left empty as Model spans down
        & 3-Person & 4-Person & 5-Person & Avg(1-5) & HPS & Action-S & Action-C \\
        \hline
        \multicolumn{8}{c}{\cellcolor{lightyellow}\textbf{Task 4}: Text-to-Image Multi-Human Generation} \\ 
        
        \hline % Horizontal rule below the header

        % --- Example Data Rows ---
        % You will replace these lines with your actual data
        SD-1.5 & 25.0 & 12.8 & 7.4 & 26.6 & 24.8 & 84.5 & 73.3 \\
        RV-1.5 & 52.5 & 20.3 & 11.5 & 43.5 & 26.9 & 88.3 & 76.0 \\
        SDXL & 44.0 & 30.7 & 23.5 & 43.3 & 26.9 & 87.9 & 79.1 \\
        SD3.5 & 61.0 & \underline{45.6} & 28.8 & \underline{56.1} & 27.8 & \textbf{95.7} & \textbf{85.0} \\
        OmniGen & \underline{64.0} & 29.0 & \underline{33.2} & 53.8 & \textbf{28.7} & \underline{93.2} & 82.5 \\
        Flux-Dev & \textbf{66.9} & \textbf{57.0} & \textbf{46.4} & \textbf{63.9} & \underline{28.2} & 92.6 & \underline{83.0} \\
        % --- End Example Data Rows ---

        \hline % Bottom horizontal rule
    \end{tabular}
    \caption{Benchmaking Foundational Text-to-Image models on generating multiple people.} % Add your table caption
    \label{tab:t2i_gen} % Add a label for cross-referencing
\end{table}

\subsection{Scope for Improvement}
\vspace{-5pt}
From the analysis in this Section and in Appendix~\ref{sec:appendixquant},\ref{sec:appendixqual}, we uncover several limitations of current approaches performing Multi-Human Generation. \textbf{First}, we notice that without regional priors, the open-source models performing Tasks 1, 2, 4 are lacking in terms of the person count and complex actions. Essentially, this means that multi-human generation is significantly hindered because the base model itself (from task 4) isn't able to generate an accurate number of human faces in a scene within the text described action. \textbf{Second}, we notice that none of the methods (with OR without regional priors) consistently pass the eye test in generating images of the correct number of people while maintaining a high ID similarity, and sufficient action scores. This balance is essential for our proposed task, and remains an open challenge. \textbf{Third}, in Appendix~\ref{sec:appendixquant}, we observe that several methods contain implicit biases over age, racial profile and gender. After uncovering these biases, we hope that the community strives to reduce them using insights from our benchmark.

\section{Conclusion}
\vspace{- 0.5 em}
This paper introduces MultiHuman-Testbench, the first comprehensive benchmark for subject-driven multi-human image generation. We contribute a carefully curated dataset of 1,800 testing samples with balanced demographic representation, a multi-faceted suite of metrics capturing count accuracy, identity preservation, visual quality, and action consistency. We also propose training-free enhancements to unified human generation models through Unified Regional Isolation and Implicit Assignment. Through extensive evaluation of approximately 30 models across four distinct tasks, we reveal that current state-of-the-art methods exhibit significant limitations. Even the best-performing models struggle with accurate human counts and preserving individual identities without subject blending artifacts. Our analysis highlights substantial opportunities for future research in achieving robust identity preservation while maintaining natural pose diversity. We believe that this benchmark can facilitate collaborative efforts to address the challenging problem of Multi-Human generation.

\bibliographystyle{plain}
\bibliography{main}
% checklist

%\clearpage

%\input{checklist}

\clearpage

\appendix
\appendixpage
\counterwithin{figure}{section}
\counterwithin{table}{section}
%-------------------------------------------------
\section{Contents}
\label{sec:SuppleIntro}

This appendix provides supplementary material to accompany the main paper. It includes a detailed breakdown of the data distribution used, further specifics on our proposed methodology, implementation details for baselines and our approach, extended quantitative and qualitative results including ablation studies and failure cases, and a discussion on the societal impact of our work. The following sections detail these aspects:

\vspace{1em} % Adds a little vertical space

\begin{itemize}[leftmargin=2em, labelwidth=!, labelsep=0.5em, itemsep=0.5ex,  align=left]
    \item[Section~\ref{sec:SuppleIntro}:] Contents
    \item[Section~\ref{sec:appendixdata}:] Data Distribution
    \item[Section~\ref{sec:appendixmethod}:] Additional Details on Proposed Approach
        \begin{itemize}[leftmargin=1.5em, labelwidth=!, labelsep=0.5em, itemsep=0.2ex, align=left]
            \item[Subsection~\ref{sec:appendixIRA}:] Implicit Region Assignment
        \end{itemize}
    \item[Section~\ref{sec:appendiximpl}:] Baselines and Implementation Details
    \item[Section~\ref{sec:appendixquant}:] Additional Quantitative Results
        \begin{itemize}[leftmargin=1.5em, labelwidth=!, labelsep=0.5em, itemsep=0.2ex, align=left]
            \item[Subsection~\ref{sec:numpeople}:] Performance across varying number of people
            \item[Subsection~\ref{sec:bias}:] Measuring Bias in Multi-Human Generation
            \item[Subsection~\ref{sec:pose}:] Effect of Our Pose Priors
        \end{itemize}
    \item[Section~\ref{sec:appendixqual}:] Additional Qualitative Results
        \begin{itemize}[leftmargin=1.5em, labelwidth=!, labelsep=0.5em, itemsep=0.2ex, align=left]
            \item[Subsection~\ref{sec:mhomni}:] Improvements from MH-OmniGen
            \item[Subsection~\ref{sec:regional}:] Qualitative results for Task 2: Regional Priors
            \item[Subsection~\ref{sec:task4}:] Qualitative results for Task 4: Text-to-Image generation
            %\item[Subsection~\ref{sec:limitations}:] Qualitative Failure Cases
        \end{itemize}
    \item[Section~\ref{sec:societal}:] Societal Impact
\end{itemize}

\section{Data Distribution}
\label{sec:appendixdata}
In this Section, we provide a detailed description of the sampled faces. This is supplementing Figure~\ref{fig:distribution} in the main text.

\begin{table}[htbp] % [h]ere, [t]op, [b]ottom, [p]age of floats
    \centering
    % --- SIUNITX SETUP ---
    % This setup will now format the numbers you provide
    % (which are already percentages) to 2 decimal places.
    \sisetup{
        round-mode=places,      % Round to a specific number of decimal places
        round-precision=2,      % Round to 2 decimal places
        output-decimal-marker={.} % Ensure dot is used as decimal marker
    }
    % The 'S' column will now just align the numbers.
    % 'table-format=2.2' means up to 2 digits before the decimal, 2 after.
    % We add the '%' symbol to the column header.
    \begin{tabular}{llS[table-format=2.2]}
        \toprule
        \textbf{Attribute} & \textbf{Category} & {\textbf{Percentage (\%)}} \\ % Added (%) to header
        \midrule
        \multirow{3}{*}{Age} & Young Adult (16-35) & 43.17 \\
                             & Middle Age (36-60)  & 42.72 \\
                             & Aged (60+)          & 14.11 \\
        \cmidrule(lr){1-3}
        \multirow{2}{*}{Gender} & Male   & 49.24 \\ % 49.297 rounded
                                & Female & 50.76 \\ % 50.702 rounded
        \cmidrule(lr){1-3}
        \multirow{2}{*}{Status} & Anonymous  & 81.48 \\ % 81.477 rounded
                                          & Celebrity & 18.52 \\ % 18.522 rounded
        \cmidrule(lr){1-3}
        \multirow{2}{*}{Data} & Real  & 72.41 \\ % 72.414 rounded
                                   & Synthetic & 27.59 \\ % 27.585 rounded (standard rounding up)
        \cmidrule(lr){1-3}
        \multirow{6}{*}{Ethnicity} & White           & 16.52 \\ % 16.540 rounded
                                   & Black           & 15.75 \\ % 15.855 rounded (standard rounding up)
                                   & South Asian     & 16.64 \\ % 16.522 rounded
                                   & East Asian      & 16.72 \\ % 16.864 rounded
                                   & Hispanic/Latin  & 16.73 \\ % 16.666 rounded
                                   & Middle Eastern/Other European  & 17.64 \\ % 17.549 rounded
        \bottomrule
    \end{tabular}
\caption{Data Distribution by Attribute. All the labels are obtained by Gemini-Flash-2.0.}
\label{tab:data_distribution_manual_percent_2dp}
\end{table}

\section{Additional Details on Proposed Approach}
\label{sec:appendixmethod}

In this Section, we provide detailed explanation of our proposed Implicit Region Assignment strategy behind the MH-Omnigen and MH-IR-Diffusion methods. This is an extension of Section~\ref{Proposed Approach: Enhancing Existing Methods}.

\subsection{Implicit Region Assignment}
\label{sec:appendixIRA}

To construct the attention mask $\mathbf{A_{iso}}$ described in Section~\ref{Proposed Approach: Enhancing Existing Methods}, we need a region of interest (ROI) set $\mathcal{R}_k \subseteq \mathcal{D}_{\text{latent}}$ for every reference image $I_k$. To facilitate the generation of multi-human images in the wild, we propose an implicit region assignment strategy that utilizes intermediate attention scores and Hungarian matching to assign each reference image to a selected region-of-interest.

\textbf{MH-Omnigen:} For unified architectures, we propose a two-stage process to determine ROIs implicitly from the model's own understanding and the intermediate latent representation $\mathbf{O}_{int,t}$. This process involves first identifying areas in the latent space which have a high self-attention overlap with each reference image. Next, we segment the estimated image from an intermediate timestep in and assign these segments to the reference images using hungarian matching.

% --- ALGORITHM 1 ---
\begin{algorithm}[htp] % Using [H] to suggest "here" placement if desired
\caption{Find attention-based similarity maps for Reference Images}
\label{alg:attention_overlap_probe_tokens_v9_direct_sep_after_outputs}
\begin{algorithmic}[1]
\renewcommand{\algorithmicrequire}{\textbf{Inputs:}}
\renewcommand{\algorithmicensure}{\textbf{Outputs:}}
\Require
    \Statex \hspace*{\algorithmicindent} \begin{minipage}[t]{\dimexpr\linewidth-\algorithmicindent\relax}
    \begin{itemize} \itemsep0pt \parskip0pt \parsep0pt
        \item[$\bullet$] Unified Multimodal Diffusion Model $U$.
        \item[$\bullet$] Intermediate latent $\mathbf{O}_{int,t} \in \mathbb{R}^{D \times D}$ at timestep $t$.
        \item[$\bullet$] Set of $K$ input reference images: $\{I_1, \dots, I_K\}$.
        \item[$\bullet$] Image token index sets: $\mathcal{D}_{\text{img}, k} \subseteq \{1, \dots, L\}$ for $k=1, \dots, K$.
        \item[$\bullet$] Latent token index set: $\mathcal{D}_{\text{latent}} \subseteq \{1, \dots, L\}$.
        \item[$\bullet$] Reshape map $\mathcal{M}_{\text{reshape}}: \mathbb{R}^{|\mathcal{D}_{\text{latent}}|} \to \mathbb{R}^{D \times D}$, where $|\mathcal{D}_{\text{latent}}| = D \times D$.
        \item[$\bullet$] Set of $H$ layers to probe: $\mathcal{L} = \{l_1, \dots, l_H\}$.
    \end{itemize}
    \end{minipage}
\Ensure
    \Statex \hspace*{\algorithmicindent} \begin{minipage}[t]{\dimexpr\linewidth-\algorithmicindent\relax}
    \begin{itemize} \itemsep0pt \parskip0pt \parsep0pt
        \item[$\bullet$] Set of $K$ similarity maps for every reference image: $\{\mathbf{S}_k \mid k=1, \dots, K\}$, each $\mathbf{S}_k \in \mathbb{R}^{D \times D}$.
    \end{itemize}
    \end{minipage}
\Statex
\Statex \hspace*{\algorithmicindent}\rule{\dimexpr\linewidth-\algorithmicindent\relax}{0.4pt}
\Statex
\State \Comment{Initialize similarity maps}
\For {$k = 1, \dots, K$}
    \State Initialize $\mathbf{S}_k$ as a $D \times D$ zero matrix.
    \State $\mathbf{S}_k[p,q] = 0$ for all $(p,q)$.
\EndFor
\State
\State \Comment{Iterate through layers}
\For {$l$ in $\mathcal{L}$}
    \State \Comment{Get attention map for layer $l$ at timestep $t$}
    \State Let $\mathbf{P}^{(l)} \in \mathbb{R}^{L \times L}$ be the self-attention map from layer $l$.
    \State $\mathbf{P}^{(l)}_{ij}$ is attention from query token $i$ to key/value token $j$, for $i, j \in \{1, \dots, L\}$.
    \State \Comment{Iterate through all reference images $I_k$}
    \For {$k = 1, \dots, K$}
        \State \Comment{Slice attention maps for latent and aggregate over tokens corresponding to $I_k$}
        \State Define flat vector $\mathbf{V}_{k,l} \in \mathbb{R}^{|\mathcal{D}_{\text{latent}}|}$.
        \State For each $i \in \mathcal{D}_{\text{latent}}$, $\mathbf{V}_{k,l}[i] = \sum_{j \in \mathcal{D}_{\text{img}, k}} \mathbf{P}^{(l)}_{ij}$.
        \State \Comment{Reshape $\mathbf{V}_{k,l}$ into spatial map $\mathbf{SM}_{k,l} \in \mathbb{R}^{D \times D}$}
        \State $\mathbf{SM}_{k,l} = \mathcal{M}_{\text{reshape}}(\mathbf{V}_{k,l})$.
        \State \Comment{Accumulate $\mathbf{SM}_{k,l}$ into $\mathbf{S}_k$}
        \State $\mathbf{S}_k \gets \mathbf{S}_k + \mathbf{SM}_{k,l}$ \Comment{Element-wise addition}
    \EndFor
\EndFor
\State
\State \Return $\{\mathbf{S}_k \mid k=1, \dots, K\}$
\end{algorithmic}
\end{algorithm}

The first stage, detailed in Algorithm~\ref{alg:attention_overlap_probe_tokens_v9_direct_sep_after_outputs}, computes attention-based similarity maps $\{\mathbf{S}_k\}_{k=1}^K$. For each reference image $I_k$, this algorithm probes the layers $\mathcal{L}$ of the unified multimodal diffusion model $U$. It aggregates the attention from latent tokens $i \in \mathcal{D}_{\text{latent}}$ to the image tokens $j \in \mathcal{D}_{\text{img}, k}$ corresponding to $I_k$. This aggregated attention is reshaped via $\mathcal{M}_{\text{reshape}}$ into a $D \times D$ spatial map $\mathbf{SM}_{k,l}$ for each layer $l$. These layer-specific maps are then accumulated to form the final similarity map $\mathbf{S}_k \in \mathbb{R}^{D \times D}$. Each $\mathbf{S}_k$ thus highlights regions in the $D \times D$ latent space which exhibit strong attention probability with the $k$-th reference image.

The second stage, outlined in Algorithm~\ref{alg:match_generated_to_reference_v2_vae}, uses the similarity maps $\mathbf{S}_k$ (generated by Algorithm~\ref{alg:attention_overlap_probe_tokens_v9_direct_sep_after_outputs}) to derive the final binary segmentation maps $\mathcal{R}_k \in \{0,1\}^{D \times D}$ which represent the ROIs.
This algorithm takes as input the predicted latent at $t=0$, $\hat{\mathbf{O}}_{\text{int},0}$ and outputs a set of $K$ assigned binary segmentation maps $\{\mathcal{R}_k \in \{0,1\}^{D \times D}\}_{k=1}^K$.

We first generate the estimated image from the predicted latent $\hat{\mathbf{O}}_{\text{int},0}$ by passing through the VAE Decoder $D_{VAE}$. This is then segmented using the Segment Anything 2~\cite{ravi2024sam} model, denoted $SAM$, to produce initial segmentation masks $\{\mathbf{M}_{seg,j}\}$.
These masks are subsequently refined by Non-Maximum Suppression (NMS) to yield $Q$ binary segmentation maps $\{\mathbf{G}_q\}_{q=1}^Q$. These maps $\mathbf{G}_q$ are transformed to $D \times D$ resolution required for comparison with the similarity maps $\mathbf{S}_k$, and represent the candidate regions within the decoded image. If no regions survive NMS ($Q=0$), the algorithm initializes all output maps $\mathcal{R}_k$ to one and terminates.

% --- ALGORITHM 2 (Modified: no explicit symbol for estimated image) ---
\begin{algorithm}[htp] % Using [H] to suggest "here" placement
\caption{Assign Regions for each Reference Image}
\label{alg:match_generated_to_reference_v2_vae} % Label remains the same or update as needed
\begin{algorithmic}[1]
\renewcommand{\algorithmicrequire}{\textbf{Inputs:}}
\renewcommand{\algorithmicensure}{\textbf{Outputs:}}
\Require
    \Statex \hspace*{\algorithmicindent} \begin{minipage}[t]{\dimexpr\linewidth-\algorithmicindent\relax}
    \begin{itemize} \itemsep0pt \parskip0pt \parsep0pt
        \item[$\bullet$] Predicted latent at $t=0$: $\hat{\mathbf{O}}_{\text{int},0} \in \mathbb{R}^{dim_{lat}}$.
        \item[$\bullet$] VAE Decoder $D_{VAE}$.
        \item[$\bullet$] Segmentation Model $SAM$.
        \item[$\bullet$] NMS threshold $\theta_{NMS}$.
        \item[$\bullet$] Set of $K$ similarity maps $\{\mathbf{S}_k \in \mathbb{R}^{D \times D}\}_{k=1}^K$ (from Alg.~\ref{alg:attention_overlap_probe_tokens_v9_direct_sep_after_outputs}).
        \item[$\bullet$] Number of input reference images $K$.
    \end{itemize}
    \end{minipage}
\Ensure
    \Statex \hspace*{\algorithmicindent} \begin{minipage}[t]{\dimexpr\linewidth-\algorithmicindent\relax}
    \begin{itemize} \itemsep0pt \parskip0pt \parsep0pt
        \item[$\bullet$] Set of $K$ assigned generated segmentation maps $\{\mathcal{R}_k \in \{0,1\}^{D \times D}\}_{k=1}^K$.
    \end{itemize}
    \end{minipage}
\Statex
\Statex \hspace*{\algorithmicindent}\rule{\dimexpr\linewidth-\algorithmicindent\relax}{0.4pt}
\Statex
\State \Comment{Step 1: Generate Segmentation Maps from $\hat{\mathbf{O}}_{\text{int},0}$}

\State $\{\mathbf{M}_{seg,j}\}_{j=1}^{Q'} \gets SAM(D_{VAE}(\hat{\mathbf{O}}_{\text{int},0}))$. \Comment{Segment VAE decoder output.}
\State $\{\mathbf{G}_q \in \{0,1\}^{D \times D}\}_{q=1}^Q \gets NMS(\{\mathbf{M}_{seg,j}\}_{j=1}^{Q'}, \theta_{NMS})$. \Comment{$Q$ maps post-NMS, at $D \times D$ res.}

\State \Comment{Step 2: Construct Cost Matrix $\mathbf{C} \in \mathbb{R}^{K \times Q}$}
\For {$k = 1, \dots, K$}
    \For {$q = 1, \dots, Q$}
        \State $\mathbf{C}[k,q] = - \sum_{p=1}^{D} \sum_{r=1}^{D} (\mathbf{S}_k[p,r] \cdot \mathbf{G}_q[p,r])$. \Comment{Negative overlap score.}
    \EndFor
\EndFor
\State
\State \Comment{Step 3: Apply Hungarian Algorithm}
\State $\mathcal{A} \gets Hungarian(\mathbf{C})$. \Comment{$\mathcal{A}$ is set of optimal $(k,q)$ pairs.}
\State
\State \Comment{Step 4: Formulate Final Output $\mathcal{R}_k$}
\State Initialize $\mathcal{R}_k$ as $D \times D$ ones matrix for $k=1, \dots, K$.
\For {each assignment $(k^*, q^*) \in \mathcal{A}$}
    \State $\mathcal{R}_{k^*} \gets \mathbf{G}_{q^*}$.
\EndFor
\State
\State \Return $\{\mathcal{R}_k\}_{k=1}^K$.
\end{algorithmic}
\end{algorithm}

If $Q > 0$, a $K \times Q$ cost matrix $\mathbf{C}$ is constructed to evaluate the compatibility between each reference image $I_k$ (via its similarity map $\mathbf{S}_k$) and each generated segmentation map $\mathbf{G}_q$.
We compute the cost of assigning generated region $\mathbf{G}_q$ to reference image $I_k$, $\mathbf{C}[k,q]$, from the overlap between $\mathbf{G}_q$ and the similarity map $\mathbf{S}_k$ (Algorithm~\ref{alg:match_generated_to_reference_v2_vae}, Line 14).
The Hungarian algorithm~\cite{Kuhn1955} is then applied to minimize the cost matrix $\mathbf{C}$. This yields a set $\mathcal{A}$ of optimal assignment pairs $(k,q)$, where each reference image $k$ is matched to at most one generated map $q$.

Finally, the output maps $\{\mathcal{R}_k\}$ are first initialized to $D \times D$ ones matrices. For each optimal assignment $(k^*, q^*) \in \mathcal{A}$, the corresponding generated segmentation map $\mathbf{G}_{q^*}$ is assigned as the output map $\mathcal{R}_{k^*}$ for the reference image $I_{k^*}$. If a reference image $I_k$ is not part of any assignment in $\mathcal{A}$ (e.g., if $K > Q$), its $\mathcal{R}_k$ remains a ones map. 

Once we find the region assignments $\{\mathcal{R}_k\}_{k=1}^K$, we apply the Unified Isolated Attention approach explained in Section~\ref{Proposed Approach: Enhancing Existing Methods} to generate the final image.

\textbf{MH-IR-Diffusion:} For models like IR-Diffusion that involve generating an initial image $I_{gen}$, followed by a final image with specific identities, we can determine regions $\mathcal{R}_k$ by directly matching facial identity cues. This method uses ArcFace embeddings to associate faces segmented from $I_{gen}$ with the $K$ input reference images $I_k$. The segmentation mask of an assigned face in $I_{gen}$ serves as the ROI $\mathcal{R}_k$.

% --- ALGORITHM for MH-IR-Diffusion based Region Assignment ---
\begin{algorithm}[h] % Using [H] to suggest "here" placement
\caption{Assign Regions for MH-IR-Diffusion using ArcFace Embeddings}
\label{alg:mh_ir_diffusion_arcface_assignment}
\begin{algorithmic}[1]
\renewcommand{\algorithmicrequire}{\textbf{Inputs:}}
\renewcommand{\algorithmicensure}{\textbf{Outputs:}}
\Require
    \Statex \hspace*{\algorithmicindent} \begin{minipage}[t]{\dimexpr\linewidth-\algorithmicindent\relax}
    \begin{itemize} \itemsep0pt \parskip0pt \parsep0pt
        \item[$\bullet$] Generated image $I_{gen}$ ($H \times W$).
        \item[$\bullet$] Set of $K$ generated reference images $\{I_k\}_{k=1}^K$.
        \item[$\bullet$] Segmentation Model $SAM$ (for face segmentation).
        \item[$\bullet$] NMS threshold $\theta_{NMS}$.
        \item[$\bullet$] ArcFace embedding function $ArcFace$.

    \end{itemize}
    \end{minipage}
\Ensure
    \Statex \hspace*{\algorithmicindent} \begin{minipage}[t]{\dimexpr\linewidth-\algorithmicindent\relax}
    \begin{itemize} \itemsep0pt \parskip0pt \parsep0pt
        \item[$\bullet$] Set of $K$ assigned face segmentation masks $\{\mathcal{R}_k \in \{0,1\}^{H \times W}\}_{k=1}^K$.
    \end{itemize}
    \end{minipage}
\Statex
\Statex \hspace*{\algorithmicindent}\rule{\dimexpr\linewidth-\algorithmicindent\relax}{0.4pt}
\Statex
\State \Comment{Step 1: Segment faces in $I_{gen}$ and compute their ArcFace embeddings}
\State $\{\mathbf{M}_{seg,j}\}_{j=1}^{Q'} \gets SAM(I_{gen})$.
\State $\{\mathbf{G}_q \in \{0,1\}^{H \times W}\}_{q=1}^Q \gets NMS(\{\mathbf{M}_{seg,j}\}_{j=1}^{Q'}, \theta_{NMS})$.

\For{$q = 1, \dots, Q$}
    \State $\mathbf{e}_{gen,q} \gets ArcFace(I_{gen}, \mathbf{G}_q)$. \Comment{Embedding for face in $I_{gen}$ at mask $\mathbf{G}_q$.}
\EndFor
\State

\State \Comment{Step 2: Compute ArcFace embeddings for $I_k$}
\For {$k = 1, \dots, K$}
    \State $\mathbf{e}_{ref,k} \gets ArcFace(I_k)$.
\EndFor
\State

\State \Comment{Step 3: Construct Cost Matrix $\mathbf{C} \in \mathbb{R}^{K \times Q}$}
\For {$k = 1, \dots, K$}
    \For {$q = 1, \dots, Q$}
        \State $\mathbf{C}[k,q] = 1 - \text{cosine\_similarity}(\mathbf{e}_{ref,k}, \mathbf{e}_{gen,q})$.
    \EndFor
\EndFor
\State

\State \Comment{Step 4: Apply Hungarian Algorithm}
\State $\mathcal{A} \gets Hungarian(\mathbf{C})$.
\State

\State \Comment{Step 5: Formulate Final Output $\mathcal{R}_k$}
\State Initialize $\mathcal{R}_k$ as $H \times W$ zeros matrix for $k=1, \dots, K$.
\For {each assignment $(k^*, q^*) \in \mathcal{A}$}
    \State $\mathcal{R}_{k^*} \gets \mathbf{G}_{q^*}$.
\EndFor
\State

\State \Return $\{\mathcal{R}_k\}_{k=1}^K$.
\end{algorithmic}
\end{algorithm}

We show our approach in Algorithm~\ref{alg:mh_ir_diffusion_arcface_assignment}. Initially, $SAM$ and NMS are used to obtain $Q$ distinct face segmentation masks $\{\mathbf{G}_q\}$ from $I_{gen}$. ArcFace embeddings are then computed for these generated face regions $\{\mathbf{e}_{gen,q}\}$ and for the reference images $\{\mathbf{e}_{ref,k}\}$. A cost matrix $\mathbf{C}$ is built using the cosine dissimilarity ($1 - \text{cosine\_similarity}$) between these embeddings. The Hungarian algorithm then finds the optimal assignment $\mathcal{A}$ between reference images and generated faces. For each match, the corresponding mask $\mathbf{G}_{q^*}$ is designated as $\mathcal{R}_{k^*}$. These pixel-space masks $\mathcal{R}_k$ are then available for subsequent processing steps.

\section{Baselines and Implementation Details}
\label{sec:appendiximpl}

In this Section, we highlight our Implementation Details for our baselines. Every experiment was performed on an Nvidia Tesla A100 GPU.

\noindent \textbf{OmniGen.} We used the official implementation of OmniGen\footnote{\url{https://github.com/VectorSpaceLab/OmniGen}} and prompted it for multi-human with and without pose conditioning. We run the default settings of 50-step inference, with a text-guidance of 2.5. For Task-1, we set image guidance scale at 2.0, and for Task-2, we found the best results with 2.8. We implemented MH-OmniGen over this repository.

\noindent \textbf{UniPortrait.} We used the official implementation of UniPortrait\footnote{\url{https://github.com/junjiehe96/UniPortrait/tree/main}} and used all the default hyper-parameters. From the original settings, we perform a 25-step inference at a guidance scale of 7.5. For pose conditioning, UniPortrait adopts controlnet, for which we set the guidance scale at 1. 

\noindent \textbf{FastComposer.} We used the official implementation of FastComposer\footnote{\url{https://github.com/mit-han-lab/fastcomposer/tree/main}} and used all the default hyper-parameters. From the original settings, we perform a 50-step inference at a guidance scale of 5. 

\noindent \textbf{OMG.} We used the official implementation of OMG\footnote{\url{https://github.com/kongzhecn/OMG.git}}, and used the the InstantID version. As the performance was poor without ControlNet, we only report the performance with regional priors, and used all the default hyper-parameters. The inference is run in 50 steps, using a CFG scale of 3.0. The controlnet and InstantID models both were weighted at 0.8, as per original implementation. To make the performance suitable to MultiHuman, we modified their detection algorithm to match all detected humans (instead of the deafult matching with "man" and "woman").

\noindent \textbf{Regional-PuLID.} We used the official implementation of Regional-PuLID\footnote{\url{https://github.com/instantX-research/Regional-Prompting-FLUX}} and made it compatible with Multi-human generation with default box priors based on number of humans. We found best results with base ratio set to 0.3. We kept the remaining hyperparameters at default settings. These include 24 inference steps and a guidance scale of 3.5.

\noindent \textbf{LoRA/MuDI.} We used the official implementation of MuDI\footnote{\url{https://github.com/agwmon/MuDI}} and trained a single LoRA for every sample. We used the default settings for training, with 2000 steps at a learning rate of $1e^{-4}$. We train in two settings, with a single view per face and 5 views per face. During in inference, we kept the default setting with an inference of 50 steps and guidance scale of 5. We run inference with both LoRA and MuDI using the provided examples. For LoRA with pose, we use the SDXL openpose controlnet with a scale of 1.0.

\noindent \textbf{MIP-Adapter.}  
We used the official implementation of MIP-Adapter\footnote{\url{https://github.com/hqhQAQ/MIP-Adapter}}, which builds upon a pretrained IP-Adapter and SDXL model, and incorporates OpenCLIP-ViT-bigG/14 as the image encoder. We loaded the released MIP-Adapter weights into this framework to better support multi-subject generation. For multi-human generation, we used prompts from our MultiHuman-Testbench and adopted the DDIM sampler with 30 inference steps and a guidance scale of 7.5. The IP-Adapter scale was set to 0.75. For pose-based regional conditioning, we followed the ControlNet implementation\footnote{\url{https://huggingface.co/thibaud/controlnet-openpose-sdxl-1.0}}.

\noindent \textbf{IP-Adapter.}  
While the pretrained IP-Adapter provides strong performance, it is not designed to directly support multiple reference images. To address this, we adopted the MIP-Adapter framework to leverage its mechanism of weight merging within the adapter layers, without loading the MIP-Adapter weights. Instead, we retained only the pretrained IP-Adapter weights in our implementation. Sampling parameters, including 30 inference steps, guidance scale of 7.5, and IP-Adapter scale of 0.75, were set identically to those in MIP-Adapter for consistency.

\noindent \textbf{RectifID.}
We utilize the official implementation provided by the authors\footnote{\url{https://github.com/feifeiobama/RectifID}}. 
Following their setup, we adopt a modified version of Stable Diffusion 1.5 released by Perflow\footnote{\url{https://huggingface.co/hansyan/perflow-sd15-dreamshaper}}, which is pretrained on the LAION-Aesthetic-5+ dataset with a particular focus on human faces and subject-centric generation.
For the inversion process, we perform 50 sampling steps using classifier-free guidance with a guidance scale of 3.0. 
In experiments involving ControlNet, all settings are kept identical except for the use of ControlNet weights\footnote{\url{https://huggingface.co/lllyasviel/control_v11p_sd15_openpose}}.

\noindent \textbf{$\lambda$-Eclipse.} 
We employed the official implementation provided by the authors\footnote{\url{https://github.com/eclipse-t2i/lambda-eclipse-inference}}, a model designed for multi-concept personalized text-to-image generation. 
This model operates within the CLIP latent space and is specifically tailored to work with the Kandinsky v2.2~\cite{kandinsky2.2} diffusion image generator.
For inference, we employed the DDIM sampler with 50 steps and a guidance scale of 7.5. 
All experiments were conducted using the provided scripts and configurations to ensure consistency and reproducibility.

% \noindent \textbf{IP-Adapter.}  
% To ensure a fair comparison with MIP-Adapter, we used the same SDXL backbone and OpenCLIP-ViT-bigG/14 as the image encoder. Unlike MIP-Adapter, which supports multi-image input natively, the original IP-Adapter requires concatenation or fusion of reference images. Instead of using image-level fusion tricks, we adopted the MIP-Adapter framework without loading its learned weights. In this setup, multiple reference images were handled via weight merging within the adapter layers. Sampling parameters, including 30 inference steps, guidance scale of 7.5, and IP-Adapter scale of 0.75, were set identically to those in MIP-Adapter for consistency.

\section{Additional Quantitative Results}
\label{sec:appendixquant}

\begin{figure*}[h]
    \centering
    \includegraphics[width=0.9\linewidth]{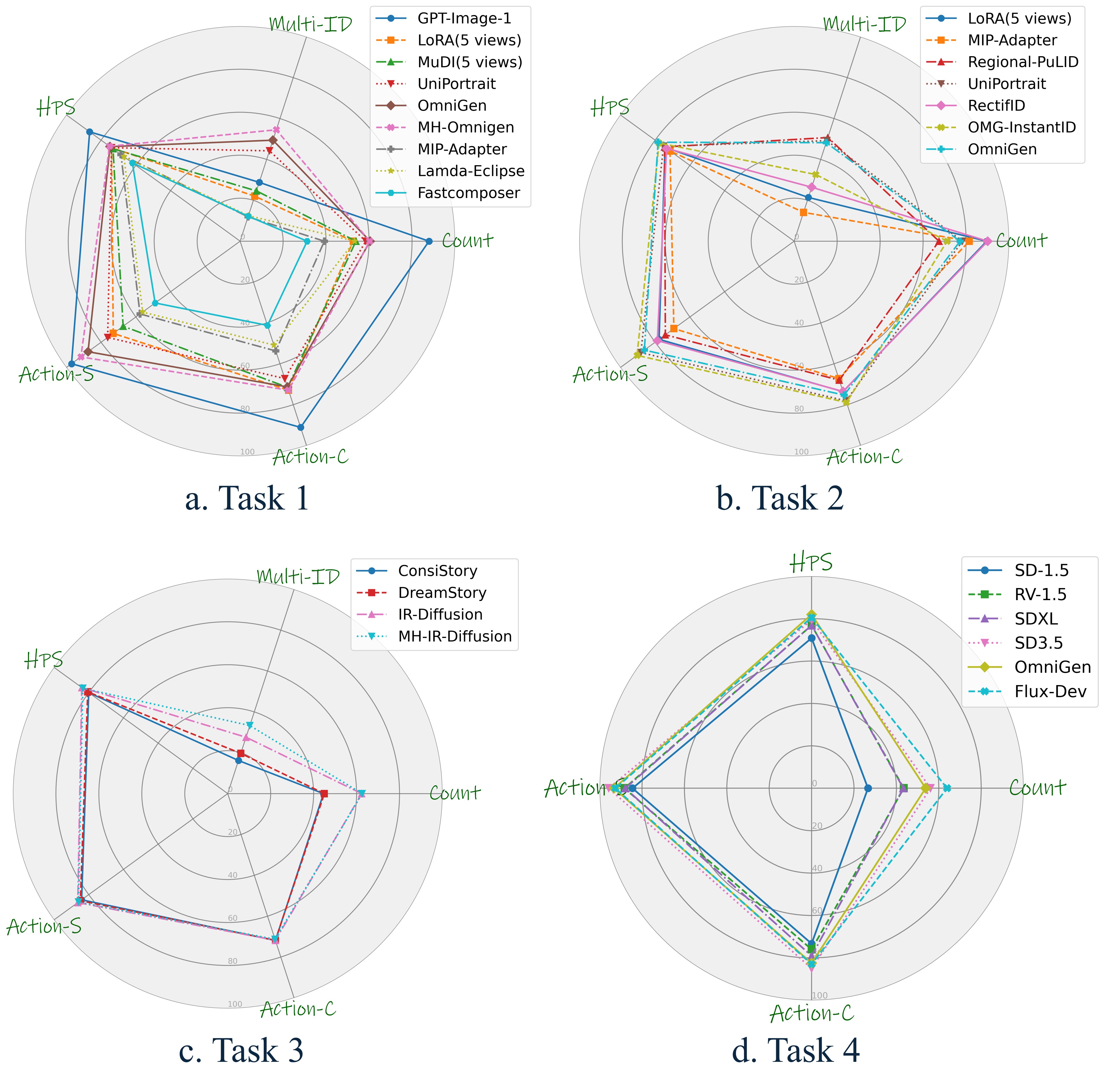}
    \vspace{- 1.2 em}
    \caption{\textbf{Radar Graphs}. Visualizing performance on Tasks 1,2,3,4 using Radar Graphs.}%, 
    \label{fig:radar}
\end{figure*}

In this Section, we provide additional Quantitative analysis on the MultiHuman Testbench, as an extension to Section~\ref{sec:benchmarking} of the paper. The main results from Section~\ref{sec:benchmarking} are summarized in the Radar graphs in Figure~\ref{fig:radar}.

\subsection{Performance across varying number of people}
\label{sec:numpeople}

In this Section, we study Multi-ID similarity scores and Person Count for methods performing Task 1, across different number of people. This is an extension to Table~\ref{tab:multihuman_noref}in the main paper. 

Summarized in Table \ref{tab:person_count}, we consistently observe that the ability to maintain ID Similarity deteriorates across all models as the number of faces in the generated image increases from two to five. This indicates a general bias where identity preservation becomes significantly more challenging with growing scene complexity. However, the magnitude and nature of this drop vary between methods.
Methods such as UniPortrait, OmniGen, and MH-OmniGen start with relatively high ID Similarity scores for one or two people, indicating strong initial performance. As the person count increases, they experience substantial absolute drops in ID Similarity. For instance, MH-OmniGen's score falls from 65.3 at two people to 38.0 at five people. Fastcomposer begins with a significantly lower ID Similarity even for just two people (15.3) and suffers the most dramatic percentage drop, falling to 5.9 at five people. GPT-Image-1 starts at a moderate ID Similarity score for two people (31.8) and exhibits the least severe relative decrease in performance as the person count increases, resulting in a score of 24.9 at five people. From the Person Count results, it is clear that GPT-Image-1 is more accurate in generating images with the correct number of people, compared to other methods, which fail often for more than three people. 

\begin{table}[htbp] % [htbp] are placement specifiers (here, top, bottom, page)
    \vspace{1.2 em}
    \renewcommand{\arraystretch}{1.5} % Increase vertical space between rows
    \fontsize{8.0pt}{6.75pt}\selectfont % Set font size
    \centering % Center the table on the page
    % Defines column format:
    % l: Left-aligned text column (for Model)
    % |: Vertical rule
    % c: Centered column (for the 5 metrics in each group)
    % Total columns: Model (l) + 5 Metrics + 5 Metrics = 11
    \begin{tabular}{l|ccccc|ccccc}
        \hline % Top horizontal rule
        % First header row: Model (spans 2 rows) and Metrics (spans 5 columns each)
        % Swapped "Multi-ID Similarity" and "Person Count"
        \multirow{2}{*}{Model} & \multicolumn{5}{c|}{Multi-ID Similarity} & \multicolumn{5}{c}{Person Count}\\
        % Second header row: specific metric names under the Metrics header
        % The first column (&) is left empty as Model spans down
        & 2 & 3 & 4 & 5 & Avg(1-5) & 2 & 3 & 4 & 5 & Avg(1-5) \\
        \hline
        \multicolumn{11}{c}{\cellcolor{lightyellow}\textbf{Task 1}: Reference-based Multi-Human Generation in the Wild} \\
        \hline % Horizontal rule below the header

        % --- Data Rows with Swapped Columns ---
        GPT-Image-1 & 31.8 & 29.5 & 27.8 & 24.9 & 28.8 & 90.7 & 91.8 & 89.5 & 75.3 & 87.9 \\

        Fastcomposer & 15.3 & 7.4 & 7.2 & 5.9 & 12.2 & 62.9 & 11.2 & 3.2 & 1.1 & 31.2 \\

        Uniportrait & 56.5 & 46.4 & 33.8 & 28.6 & 44.2 & 90.6 & 76.3 & 23.7 & 14.1 & 58.5 \\

        OmniGen & 60.8 & 52.3 & 42.2 & 35.2 & 49.4 & 88.8 & 88.0 & 23.2 & 21.6 & 60.5 \\

        MH-OmniGen & 65.3 & 60.4 & 45.1 & 38.0 & 54.5 & 91.2 & 87.5 & 22.4 & 19.7 & 60.3 \\
        % --- End Data Rows ---

        \hline % Bottom horizontal rule
    \end{tabular}
    \caption{Studying the ID similarity and Person count metrics for different number of people, for Multi-Human Generation in the Wild.} % Add your table caption
    \label{tab:person_count} % Consider updating label if it's a new distinct table
\end{table}

\subsection{Measuring Bias in Multi-Human Generation}
\label{sec:bias}

Using the labeled attributes provided with the data, we provide a study on the biases for each method in Task 1 and Task 2. We measure the single-person ID similarity and report it across various splits: Ethnicity(~\ref{tab:multihuman_ref_race_nobackbone}), age, gender and status(~\ref{tab:multihuman_ref_age_type_nobackbone}). Based on the ID-Similarity scores highlighted in red in Table \ref{tab:multihuman_ref_race_nobackbone} and Table \ref{tab:multihuman_ref_age_type_nobackbone}, biases are evident in how different models and their backbones perform. These biases are indicated by deviations from the average score for each row. A darker red shades signifies higher bias.

In Task 1, GPT-Image-1 (GPT-4o) shows a positive bias for South-Asian identities and a positive bias for Aged individuals. UniPortrait (RV1.5) exhibits positive biases favoring White and East-Asian faces while underperforming on Black individuals. Additionally, it heavily underperforms on Aged individuals by demographic type, offset by a positive bias for Young Adults. Fastcomposer (SD1.5) shows minimal racial/ethnic bias but has a light negative bias for Celeb faces. OmniGen (Phi-3) and MH-OmniGen (Phi-3) generally display less pronounced biases in Task 1, showing mostly light biases favoring White and South-Asian faces by race/ethnicity, suggesting more balanced performance compared to the rest.

% Table 1: Race/Ethnicity Bias
\begin{table}[htbp] % [htbp] are placement specifiers (here, top, bottom, page)
    \renewcommand{\arraystretch}{1.5}
    \fontsize{8.0pt}{6.75pt}\selectfont
    \centering % Center the table on the page
    \begin{tabular}{l| c | cccccc} % Defines column format: Model (l) + Backbone (c) + ID Sim (6xc) = 8
        \hline % Top horizontal rule from booktabs
        \multirow{2}{*}{Model} & \multirow{2}{*}{\parbox{1.5cm}{Backbone}} & \multicolumn{6}{c}{ID-Similarity}\\ % Multicolumn span remains 6
        \cline{3-8} % Added cline for clarity between headers
        & & White & Black & South-Asian & East-Asian & Hispanic & Middle-East \\ % Metric names remain
        \hline
        \multicolumn{8}{c}{\cellcolor{lightyellow}\textbf{Task 1}: Reference-based Multi-Human Generation in the Wild} \\ % Adjusted multicolumn span to 8
        \hline  % Middle horizontal rule from booktabs

        % --- Multi-Human Tuning-Free Rows (No regional Priors) ---
        % Data from input with coloring based on deviation
        % GPT-Image-1: Mean=27.88. Devs: 1.48(L), 0.92(N), 2.32(M), 1.68(L), 0.38(N), 0.32(N). Thresholds: N<1, L[1,2), M[2,4), D>=4.
        GPT-Image-1 & GPT-4o & \cellcolor{red!10} 26.4 & 28.8 & \cellcolor{red!25} 30.2 & \cellcolor{red!10} 26.2 & 27.5 & 28.2 \\
        \hline % Added hline to separate model blocks
        % UniPortrait: Mean=38.98. Devs: 2.12(M), 2.18(M), 0.68(N), 2.82(M), 1.48(L), 0.58(N).
        UniPortrait & RV1.5 & \cellcolor{red!25} 41.1 & \cellcolor{red!25} 36.8 & 38.3 & \cellcolor{red!25} 41.8 & \cellcolor{red!10} 37.5 & 38.4 \\
        % Fastcomposer: Mean=9.07. Devs: 0.37(N), 0.27(N), 0.23(N), 0.43(N), 0.13(N), 0.17(N). -> No shading
        Fastcomposer & SD1.5 & 8.7 & 8.8 & 9.3 & 9.5 & 9.2 & 8.9 \\
        \hline % Added hline to separate model blocks
        % OmniGen (T1): Mean=45.15. Devs: 1.15(L), 0.05(N), 1.85(L), 0.15(N), 0.35(N), 0.85(N).
        OmniGen & Phi-3 & \cellcolor{red!10} 44.0 & 45.1 & \cellcolor{red!10} 47.0 & 45.0 & 45.5 & 44.3 \\
        % MH-OmniGen (T1): Mean=49.67. Devs: 1.27(L), 0.37(N), 1.93(L), 0.33(N), 0.03(N), 0.67(N).
        MH-OmniGen & Phi-3 & \cellcolor{red!10} 48.4 & 49.3 & \cellcolor{red!10} 51.6 & 50.0 & 49.7 & 49.0 \\
        \hline

        \multicolumn{8}{c}{\cellcolor{lightyellow}\textbf{Task 2}: Reference-based Multi-Human Generation with Regional Priors} \\ % Adjusted multicolumn span to 8
        \hline % Top horizontal rule from booktabs

        % --- Multi-Human Tuning-Free Rows (With regional Priors) ---
        % Data from input with coloring based on deviation
        % UniPortrait (T2): Mean=45.62. Devs: 3.28(M), 2.02(M), 0.32(N), 1.68(L), 2.32(M), 0.32(N).
        UniPortrait & RV1.5 & \cellcolor{red!25} 48.9 & \cellcolor{red!25} 43.6 & 45.3 & \cellcolor{red!10} 47.3 & \cellcolor{red!25} 43.3 & 45.3 \\
        % RectifID (T2): Mean=18.03. Devs: 0.43(N), 1.13(L), 1.17(L), 3.07(M), 1.73(L), 0.93(N).
        RectifID & SD1.5 & 17.6 & \cellcolor{red!10} 16.9 & \cellcolor{red!10} 19.2 & \cellcolor{red!25} 21.1 & \cellcolor{red!10} 16.3 & 17.1 \\
        \hline % Added hline to separate model blocks
        % Regional-PuLID (T2): Mean=47.55. Devs: 0.15(N), 3.45(M), 0.35(N), 5.05(D), 0.25(N), 1.55(L).
        Regional-PuLID & Flux & 47.4 & \cellcolor{red!25} 44.1 & 47.9 & \cellcolor{red!40} 52.6 & 47.3 & \cellcolor{red!10} 46.0 \\
        \hline % Added hline to separate model blocks
        % OMG-InstantID (T2): Mean=26.62. Devs: 0.18(N), 2.32(M), 0.38(N), 3.98(M), 0.72(N), 1.52(L).
        OMG-InstantID & SDXL & 26.8 & \cellcolor{red!25} 24.3 & 27.0 & \cellcolor{red!25} 30.6 & 25.9 & \cellcolor{red!10} 25.1 \\
        \hline % Added hline to separate model blocks
        % OmniGen (T2): Mean=42.73. Devs: 0.03(N), 0.13(N), 2.17(M), 0.87(N), 1.33(L), 1.53(L).
        OmniGen & Phi-3 & 42.7 & 42.6 & \cellcolor{red!25} 44.9 & 43.6 & \cellcolor{red!10} 41.4 & \cellcolor{red!10} 41.2 \\
        \bottomrule % Bottom horizontal rule from bookbooks
    \end{tabular}
    \caption{Multi-Human Generation with Reference Images: Multi-Human Tuning-Free Models with ID-Similarity Metrics by Race/Ethnicity (Backbone Removed)} % Original caption
    \label{tab:multihuman_ref_race_nobackbone} % Original label
\end{table}

Turning to Task 2, where regional priors are used, the patterns of bias shift for some models. UniPortrait (RV1.5) continues to show biases favoring White and East-Asian faces and against Black and Hispanic faces, favoring female faces to male faces, and heavily favoring young adults. RectifID (SD1.5) shows a bias favoring East-Asian faces by race/ethnicity. Regional-PuLID (Flux) displays significant biases, with a strong positive bias for East-Asian individuals and a negative bias against Black faces by ethnicity. By demographic, Regional-PuLID exhibits strong biases against Males and Aged faces, while strongly favoring Female, Young Adult, and Celeb identities. OMG-InstantID (SDXL) shows a bias against Black faces and favoring East-Asian faces, and favors Young Adults. OmniGen (Phi-3) in Task 2 shows less prominent biases, with a bias favoring South-Asian faces and light biases against Hispanic and Middle-East faces by ethnicity.

% Table 2: Demographic/Type Bias
\begin{table}[htbp] % [htbp] are placement specifiers (here, top, bottom, page)
    \renewcommand{\arraystretch}{1.5}
    \fontsize{8.0pt}{6.75pt}\selectfont
    \centering % Center the table on the page
    \begin{tabular}{l| c | cc | ccc | cc} % Defines column format: Model (l), Backbone (c), 7 ID Sim (c)
        \hline % Top horizontal rule from booktabs
        \multirow{2}{*}{Model} & \multirow{2}{*}{\parbox{1.5cm}{Backbone}} & \multicolumn{7}{c}{ID-Similarity}\\ % Changed multicolumn span from 5 to 7
        \cline{3-9}
        & & Male & Female & Young Adult & Middle Aged & Aged & Celeb & Anonymous \\ % Added Celeb and Anonymous
        \hline
        \multicolumn{9}{c}{\cellcolor{lightyellow}\textbf{Task 1}: Reference-based Multi-Human Generation in the Wild} \\ % Adjusted multicolumn span from 7 to 9
        \hline  % Middle horizontal rule from booktabs

        % --- Multi-Human Tuning-Free Rows (No regional Priors) ---
        % Data from input with coloring based on deviation - 7 columns for Demographic/Type
        % GPT-Image-1: Mean=27.97. Devs: |29.8-27.97|=1.83(L), |26.0-27.97|=1.97(L), |26.0-27.97|=1.97(L), |28.8-27.97|=0.83(N), |30.8-27.97|=2.83(M), |28.1-27.97|=0.13(N), |26.6-27.97|=1.37(L). Thresholds: N<1, L[1,2), M[2,4), D>=4.
        GPT-Image-1 & GPT-4o & \cellcolor{red!10} 29.8 & \cellcolor{red!10} 26.0 & \cellcolor{red!10} 26.0 & 28.8 & \cellcolor{red!25} 30.8 & 28.1 & \cellcolor{red!10} 26.6 \\
        \hline % Added hline
        % UniPortrait (T1): Mean=38.57. Devs: |37.5-38.57|=1.07(L), |40.5-38.57|=1.93(L), |40.7-38.57|=2.13(M), |37.7-38.57|=0.87(N), |30.7-38.57|=7.87(D), |37.6-38.57|=0.97(N), |39.3-38.57|=0.73(N).
        UniPortrait & RV1.5 & \cellcolor{red!10} 37.5 & \cellcolor{red!10} 40.5 & \cellcolor{red!25} 40.7 & 37.7 & \cellcolor{red!40} 30.7 & 37.6 & 39.3  \\
        % Fastcomposer (T1): Mean=8.93. Devs: |9.0-8.93|=0.07(N), |9.1-8.93|=0.17(N), |9.2-8.93|=0.27(N), |9.1-8.93|=0.17(N), |8.5-8.93|=0.43(N), |7.1-8.93|=1.83(L), |9.5-8.93|=0.57(N).
        Fastcomposer & SD1.5 & 9.0 & 9.1 & 9.2 & 9.1 & 8.5 & \cellcolor{red!10} 7.1 & 9.5 \\
        \hline % Added hline
        % OmniGen (T1): Mean=45.03. Devs: |44.9-45.03|=0.13(N), |45.4-45.03|=0.37(N), |44.3-45.03|=0.73(N), |46.0-45.03|=0.97(N), |45.4-45.03|=0.37(N), |45.5-45.03|=0.47(N), |43.7-45.03|=1.33(L).
        OmniGen & Phi-3 & 44.9 & 45.4 & 44.3 & 46.0 & 45.4 & 45.5 & \cellcolor{red!10} 43.7 \\
        % MH-OmniGen (T1): Mean=49.79. Devs: |49.5-49.79|=0.29(N), |49.8-49.79|=0.01(N), |49.0-49.79|=0.79(N), |49.9-49.79|=0.11(N), |50.7-49.79|=0.91(N), |48.7-49.79|=1.09(L), |49.9-49.79|=0.11(N).
        MH-OmniGen & Phi-3  & 49.5 & 49.8 & 49.0 & 49.9 & 50.7 & \cellcolor{red!10} 48.7 & 49.9 \\
        \hline

        \multicolumn{9}{c}{\cellcolor{lightyellow}\textbf{Task 2}: Reference-based Multi-Human Generation with Regional Priors} \\ % Adjusted multicolumn span from 7 to 9
        \hline % Top horizontal rule from booktabs

        % --- Multi-Human Tuning-Free Rows (With regional Priors) ---
        % Data from input with coloring based on deviation - 7 columns for Demographic/Type
        % UniPortrait (T2): Mean=45.50. Devs: |43.5-45.50|=2.00(M), |47.7-45.50|=2.20(M), |48.2-45.50|=2.70(M), |44.0-45.50|=1.50(L), |42.9-45.50|=2.60(M), |46.9-45.50|=1.40(L), |45.3-45.50|=0.20(N).
        UniPortrait & RV1.5 & \cellcolor{red!25} 43.5 & \cellcolor{red!25} 47.7 & \cellcolor{red!25} 48.2 & \cellcolor{red!10} 44.0 & \cellcolor{red!25} 42.9 & \cellcolor{red!10} 46.9 & 45.3 \\
        % RectifID (T2): Mean=17.97. Devs: |17.4-17.97|=0.57(N), |18.6-17.97|=0.63(N), |19.1-17.97|=1.13(L), |17.4-17.97|=0.57(N), |16.1-17.97|=1.87(L), |19.5-17.97|=1.53(L), |17.7-17.97|=0.27(N).
        RectifID & SD1.5 & 17.4 & 18.6 & \cellcolor{red!10} 19.1 & 17.4 & \cellcolor{red!10} 16.1 & \cellcolor{red!10} 19.5 & 17.7 \\
        \hline % Added hline
        % Regional-PuLID (T2): Mean=47.26. Devs: |42.3-47.26|=4.96(D), |52.5-47.26|=5.24(D), |51.7-47.26|=4.44(D), |45.4-47.26|=1.86(L), |41.5-47.26|=5.76(D), |50.5-47.26|=3.24(M), |46.9-47.26|=0.36(N).
        Regional-PuLID & Flux & \cellcolor{red!40} 42.3 & \cellcolor{red!40} 52.5 & \cellcolor{red!40} 51.7 & \cellcolor{red!10} 45.4 & \cellcolor{red!40} 41.5 & \cellcolor{red!25} 50.5 & 46.9 \\
        \hline % Added hline
        % OMG-InstantID (T2): Mean=26.66. Devs: |25.0-26.66|=1.66(L), |28.2-26.66|=1.54(L), |28.7-26.66|=2.04(M), |25.1-26.66|=1.56(L), |24.9-26.66|=1.76(L), |24.9-26.66|=1.76(L), |26.8-26.66|=0.14(N).
        OMG-InstantID & SDXL & \cellcolor{red!10} 25.0 & \cellcolor{red!10} 28.2 & \cellcolor{red!25} 28.7 & \cellcolor{red!10} 25.1 & \cellcolor{red!10} 24.9 & \cellcolor{red!10} 24.9 & 26.8 \\
        \hline % Added hline
        % OmniGen (T2): Mean=42.97. Devs: |43.6-42.97|=0.63(N), |42.1-42.97|=0.87(N), |41.8-42.97|=1.17(L), |43.2-42.97|=0.23(N), |44.6-42.97|=1.63(L), |42.6-42.97|=0.37(N), |42.9-42.97|=0.07(N).
        OmniGen & Phi-3 & 43.6 & 42.1 & \cellcolor{red!10} 41.8 & 43.2 & \cellcolor{red!10} 44.6 & 42.6 & 42.9 \\
        \bottomrule % Bottom horizontal rule from bookbooks
    \end{tabular}
    \caption{Multi-Human Generation with Reference Images: Multi-Human Tuning-Free Models with ID-Similarity Metrics by Demographic and Type (Backbone Removed)} % Original caption
    \label{tab:multihuman_ref_age_type_nobackbone} % Original label
\end{table}

\subsection{Effect of Our Pose Priors}
\label{sec:pose}

Table~\ref{tab:comparison_priors} shows the effect of adding our human-rectified regional pose priors. As observed, all metrics significantly improve in case of most methods. There is a slight drop in Multi-ID and Action-S for OmniGen, and a slight drop in Multi-ID in LoRA. Overall, regional pose priors can help generate significantly better results, as they make the task easier. However, this benefit comes with a severe hit to usability, which is why solving Task 1 is important.

\begin{table}[htbp] % [htbp] are placement specifiers (here, top, bottom, page)
    \renewcommand{\arraystretch}{1.5}
    \fontsize{8.0pt}{6.75pt}\selectfont
    \centering % Center the table on the page
    \begin{tabular}{l | cc | cc | cc | cc | cc} % Defines column format: Model (l), 5 Metrics (each has 2 columns)
        \hline % Top horizontal rule
        \multirow{2}{*}{Model} & \multicolumn{2}{c|}{Count} & \multicolumn{2}{c|}{Multi-ID} & \multicolumn{2}{c|}{HPS} & \multicolumn{2}{c|}{Action-S} & \multicolumn{2}{c}{Action-C} \\
        \cline{2-11} % Horizontal rule below multi-column headers, spanning metric columns
        & Task 1 & Task 2 & Task 1 & Task 2 & Task 1 & Task 2 & Task 1 & Task 2 & Task 1 & Task 2 \\
        \hline
        % Data rows comparing Task 1 and Task 2 for each common method
        LoRA(5 views) & 52.6 & \textbf{89.6} & \textbf{22.0} & 21.4 & 25.9 & \textbf{26.0} & 73.0 & \textbf{77.7} & 72.9 & \textbf{73.6} \\
        MIP-Adapter & 39.2 & \textbf{81.5} & 11.9 & \textbf{14.1} & 24.0 & \textbf{25.0} & 57.6 & \textbf{69.2} & 53.7 & \textbf{67.2} \\
        UniPortrait & 58.5 & \textbf{78.3} & 44.2 & \textbf{49.2} & 25.9 & \textbf{26.3} & 76.2 & \textbf{88.2} & 67.2 & \textbf{78.1} \\
        RectifID & 37.8 & \textbf{90.1} & 18.6 & \textbf{26.4} & 24.8 & \textbf{25.7} & 67.3 & \textbf{78.7} & 68.2 & \textbf{73.5} \\
        OmniGen & 60.5 & \textbf{77.2} & \textbf{49.4} & 48.2 & 26.2 & \textbf{27.4} & \textbf{87.5} & 86.2 & 71.3 & \textbf{75.3} \\
        % Note: GPT-Image-1, Fastcomposer, MH-OmniGen, Regional-PuLID, OMG-InstantID are not in both lists of methods.
        \bottomrule % Bottom horizontal rule
    \end{tabular}
    \caption{Comparison of Multi-Human Generation Metrics With and Without Regional Priors for Tuning-Free Models} % Table caption
    \label{tab:comparison_priors} % Label for cross-referencing
\end{table}

\section{Additional Qualitative Results}
\label{sec:appendixqual}

In this Section, we provide additional Qualitative results to supplement the ones in Section~\ref{sec:benchmarking} of the main text.

\subsection{Improvements from MH-OmniGen}
\label{sec:mhomni}

\begin{figure*}[h]
    \centering
    \includegraphics[width=0.8\linewidth]{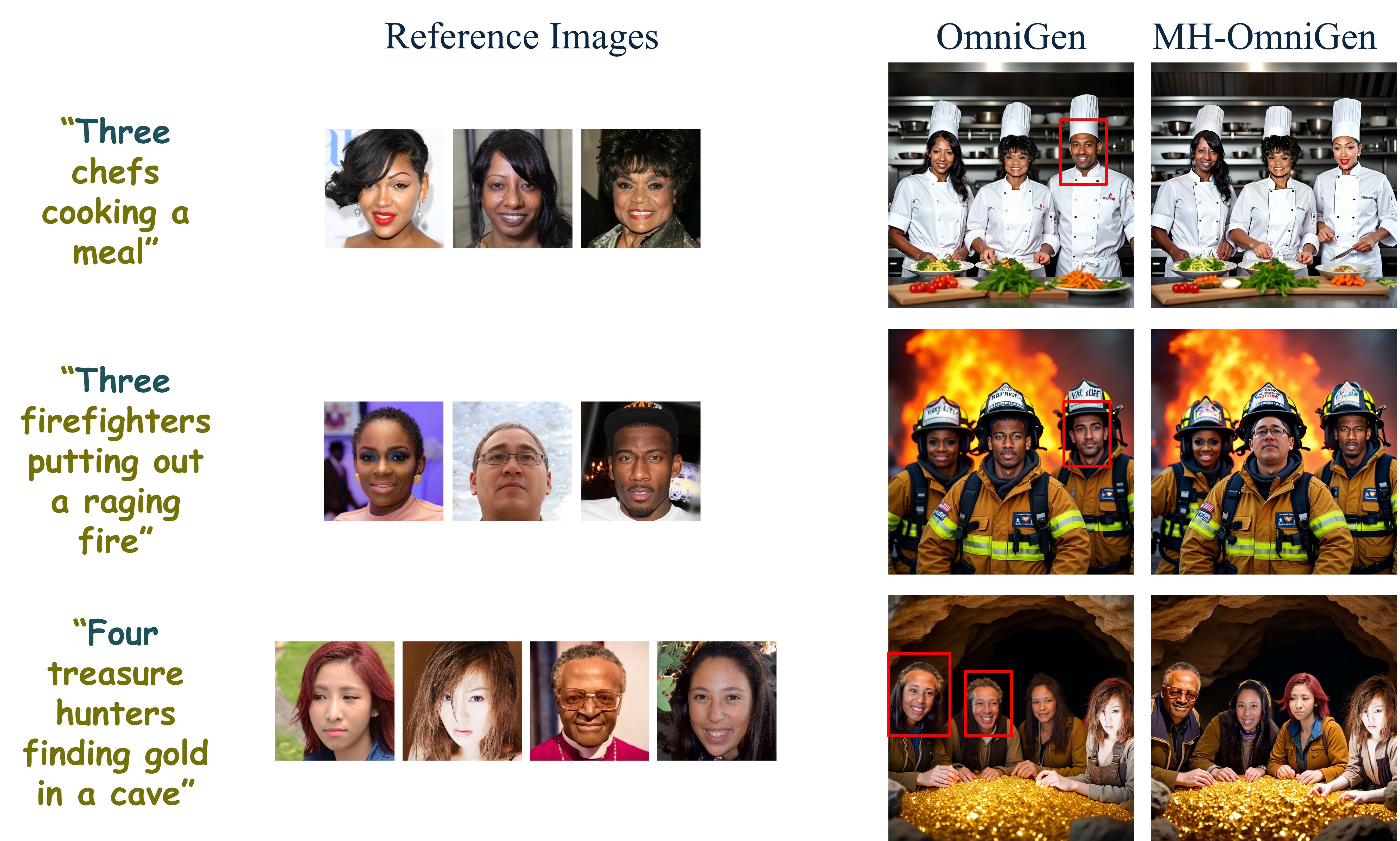}

    \caption{\textbf{Qualitative Results for MH-OmniGen v/s OmniGen on Task 1}. As observed, MH-OmniGen is able to significantly improve OmniGen by reducing ID leakage, which is highlighted with red boxes.}%, 
    \label{fig:mh_vs_omni}
\end{figure*}

Figure~\ref{fig:mh_vs_omni} shows more qualitative results on Multi-Human generation in the wild, using OmniGen and MH-OmniGen. As observed, MH-OmniGen is able to correct many instances of subject blending using our proposed Unified Region Isolation and Implicit Matching algorithm.

\subsection{Qualitative results for Task 2: Regional Priors}
\label{sec:regional}

\begin{figure*}[h]
    \centering
    \includegraphics[width=1\linewidth]{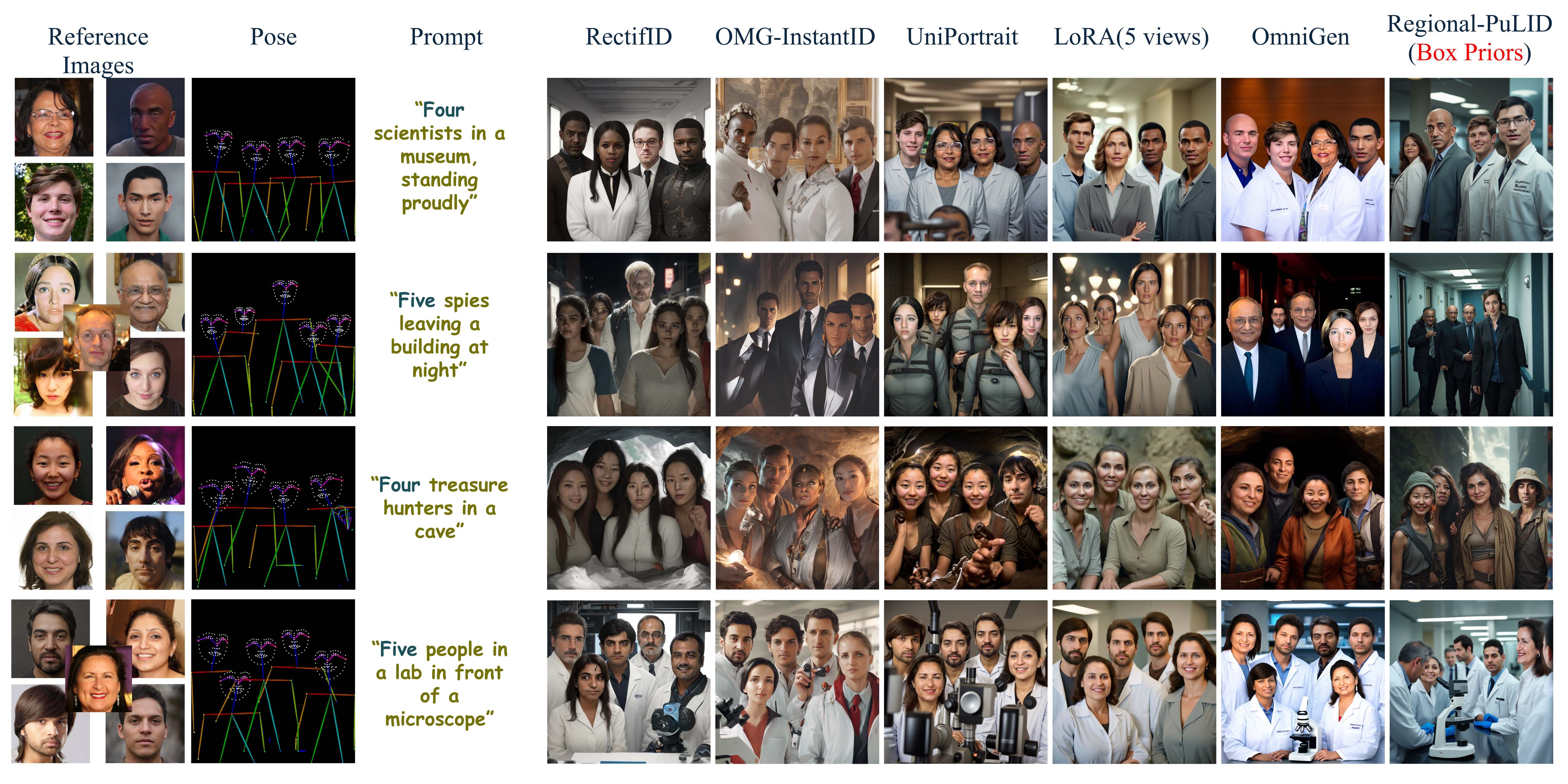}
    \vspace{- 1.2 em}
    \caption{\textbf{Qualitative Results on Multi-Human Generation with Pose conditioning}. The image shows the best performing methods: RectifID, UniPortrait, LoRA, OmniGen and Regional-PuLID(with box priors). OmniGen, UniPortrait and Regional-PuLID show best results albeit with severe limitations.}%, 
    \label{fig:pose_qual}
    \vspace{- 0.5 em}
\end{figure*}

Figure~\ref{fig:pose_qual} shows the best performing methods: RectifID, UniPortrait, LoRA, OmniGen and Regional-PuLID(with box priors) on Task 2: Reference-based Multi-Human Generation with Regional Priors. As observed in the figure, OmniGen, UniPortrait and Regional-PuLID show best results. However, it is clear that each method has severe limitations including incorrect count accuracy and underperformance on ID similarity. This underlines a huge scope for improvement.

\subsection{Qualitative results for Task 4: Text-to-Image generation}
\label{sec:task4}
\begin{figure*}[h]
    \centering
    \includegraphics[width=1\linewidth]{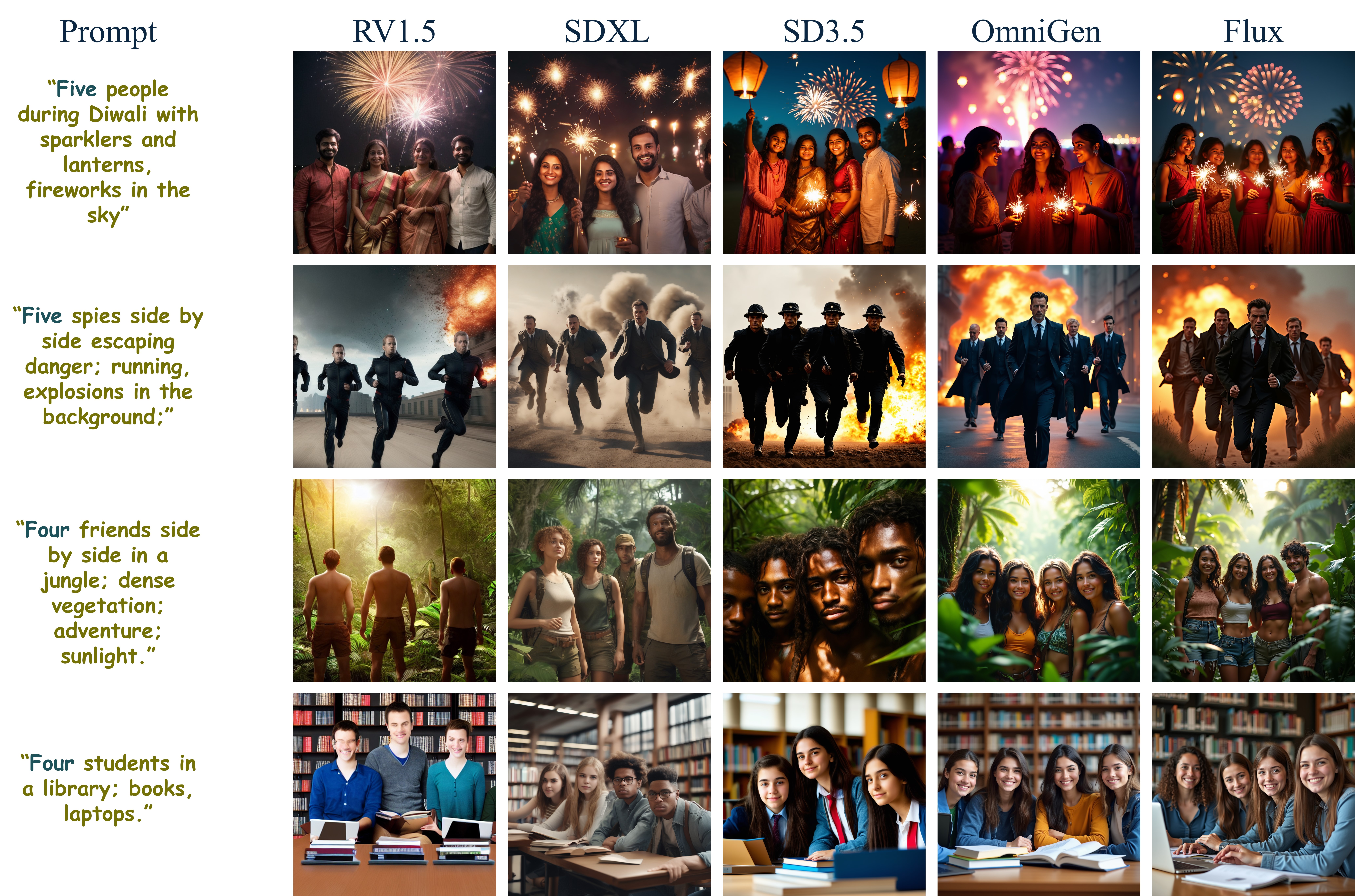}
    \vspace{- 1.2 em}
    \caption{\textbf{Qualitative Results on Multi-Human Generation for Text-to-Image models}. The image shows the best performing methods on Task 4: RV1.5, SDXL, SD3.5, OmniGen and Flux. Flux and OmniGen show best results albeit all methods show limitations in terms of count accuracy.}%, 
    \label{fig:task4}
    \vspace{- 0.5 em}
\end{figure*}

Figure~\ref{fig:task4} shows the best performing methods on Task 4: Text-to-Image Generation for Multiple Humans (with no reference images). The methods on display are RV1.5, SDXL, SD3.5, OmniGen and Flux. Flux and OmniGen show best results. However, it is clear that all methods show limitations in terms of count accuracy. Additionally, every method is susceptible to generating faces of people with similar attributes (age, race, gender). We believe that there is a heavy scope for improvement for Generation Models in this regard.

%\subsection{Qualitative Failure Cases}
%\label{sec:limitations}

\section{Societal Impact}
\label{sec:societal}

With MultiHuman-Testbench, we aim to make significant advancements in AI-driven multi-human image generation, and we anticipate substantial positive societal benefits. By encouraging the development of models which accurately depict diverse individuals across age, ethnicity, and gender while preserving their identities in complex scenes, we hope to contribute to more equitable and inclusive digital media. We envision that our benchmark can enhance creative tools for artists and developers, enrich AR/VR/XR experiences, and improve assistive technologies. Furthermore, we believe that our proposed standardized evaluation suite will accelerate research and offer clearer insights into generation model capabilities. 

However, we also recognize that this progress amplifies societal risks. The capability for highly realistic multi-human image generation increases the potential for deepfakes which could be used in misinformation campaigns or impersonation, thereby posing threats to individual privacy and societal trust. Finally, we acknowledge that the increasing sophistication of these generative tools raises concerns about job displacement in creative sectors. Hence, we request the broader community to proactively engage in developing ethical frameworks, and responsible use guidelines.

%%%%%%%%%%%%%%%%%%%%%%%%%%%%%%%%%%%%%%%%%%%%%%%%%%%%%%%%%%%%

\end{document}